\documentclass[11pt]{article}

\usepackage[normalem]{ulem}

\usepackage[final]{acl}

\usepackage{times}
\usepackage{latexsym}
\usepackage{booktabs}
\usepackage{ragged2e}
\usepackage{tabularx}

\usepackage[T1]{fontenc}

\usepackage[utf8]{inputenc}

\usepackage{microtype}

\usepackage{inconsolata}

\usepackage{graphicx}
\usepackage{subcaption}
\usepackage{tcolorbox}

\usepackage[english,bidi=default]{babel} 
\babelfont{rm}{TeXGyreTermesX} 
\babelprovide[import]{hindi}

\babelfont[hindi]{rm}[
    Path=./fonts/
]{NotoSansDevanagari-Regular.ttf}

\newcommand{\mainset}{\textsc{IndicIFEval}}
\newcommand{\transset}{\textsc{IndicIFEval-Trans}}
\newcommand{\groundset}{\textsc{IndicIFEval-Ground}}

\title{IndicIFEval: A Benchmark for Verifiable Instruction-Following Evaluation in 14 Indic Languages}

\author{
  \textbf{Thanmay Jayakumar\textsuperscript{1,2}}
  \textbf{Mohammed Safi Ur Rahman Khan\textsuperscript{1,2}}\\
  \textbf{Raj Dabre\textsuperscript{1,2}}
  \textbf{Ratish Puduppully\textsuperscript{3}},
  \textbf{Anoop Kunchukuttan\textsuperscript{1, 4}}, 
  \\
  \textsuperscript{1}Nilekani Centre at AI4Bharat,
  \textsuperscript{2}Indian Institute of Technology Madras, India, \\
  \textsuperscript{3}IT University of Copenhagen,
  \textsuperscript{4}Microsoft, India
}
\usepackage[table]{xcolor}
\usepackage{booktabs}
\definecolor{propbest}{RGB}{255, 204, 204} 
\definecolor{openbest}{RGB}{204, 229, 255} 
\definecolor{tablegray}{RGB}{243, 244, 246}
\definecolor{tableblue}{RGB}{223, 236, 254}
\definecolor{headergray}{gray}{0.90}

\begin{document}
\maketitle
\begin{abstract}
Instruction-following benchmarks remain predominantly English-centric, 
leaving a critical evaluation gap for the hundreds of millions of Indic 
language speakers. We introduce \mainset, a benchmark evaluating constrained 
generation of LLMs across 14 Indic languages using automatically verifiable, 
rule-based instructions. It comprises around 800 human-verified examples 
per language spread across two complementary subsets: \transset{}, 
translated prompts from \textit{IFEval} \cite{ifeval} carefully localized for Indic 
contexts, and \groundset{}, synthetically generated instructions grounded 
in native Indic content. We conduct a comprehensive evaluation of major open-weight and proprietary models spanning both reasoning and 
non-reasoning models. While models 
maintain strong adherence to formatting constraints, they struggle 
significantly with lexical and cross-lingual tasks -- and despite progress 
in high-resource languages, instruction-following across the broader Indic 
family lags significantly behind English. We release \mainset{} and its 
evaluation scripts to support progress on multilingual constrained generation \footnote{\url{http://github.com/ai4bharat/IndicIFEval}}.

\end{abstract}

\begin{figure*}[!h]
    \centering
    \includegraphics[width=\linewidth]{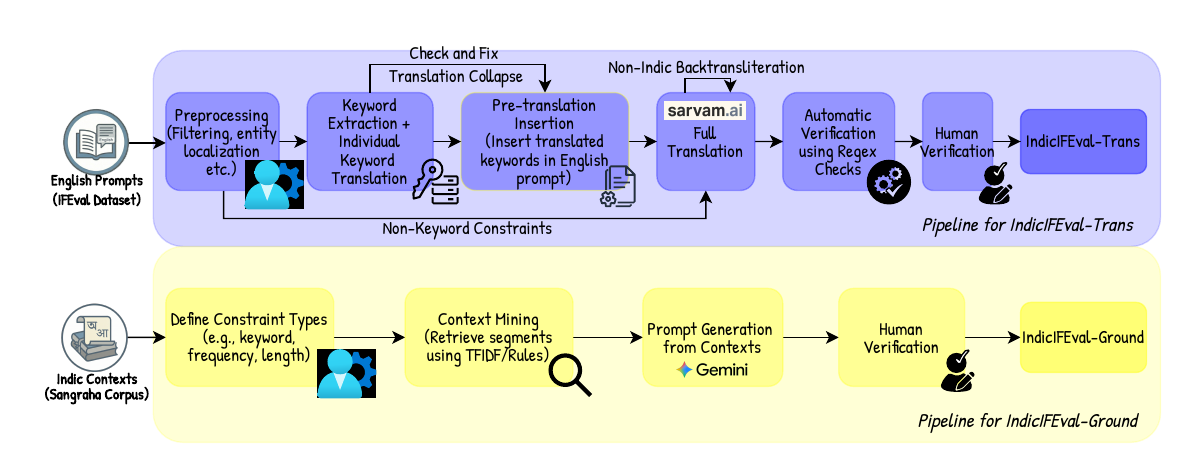}
    \caption{Overview of the dataset construction pipeline}
    \label{fig:overview_figure}
\end{figure*}

\section{Introduction}

Instruction following, the ability of a model to accurately interpret 
user intent and comply with specified instructions, is a critical aspect 
of LLM training and evaluation. This capability is commonly assessed using 
either human annotators \cite{karpinska-etal-2021-perils, chiang2024chatbot} 
or LLM-as-a-Judge approaches \cite{zheng2023judging, alpaca_eval}. However, 
human evaluation is costly and time-consuming, while LLM-based judges can 
be inconsistent and biased \cite{wang-etal-2024-large-language-models-fair, 
zheng2023judging}. A complementary paradigm using automatically verifiable 
instructions (where adherence can be determined through simple rule-based 
checks) has therefore gained traction, as exemplified by the IFEval 
benchmark \cite{ifeval}. These include instructions such as requesting 
the presence of certain words, a fixed number of sentences, or specific 
formats (e.g., JSON), all checkable via automated scripts.

At a multilingual level, this evaluation gap becomes more pronounced, as 
LLMs are often significantly weaker in non-English languages  -- not just as evaluators \cite{hada-etal-2024-large, fu-liu-2025-reliable} but also in text generation \cite{ahuja-etal-2023-mega, ustun-etal-2024-aya}.
This raises a natural question: 
\textit{In multilingual and low-resource settings, to what extent can LLMs at least 
follow simple, rule-based, verifiable constraints?}

We address this gap with \mainset, a benchmark evaluating verifiable 
instruction-following across 14 Indic languages, a linguistically 
diverse family characterized by data scarcity, morphological richness, 
and distinct script systems, making it a meaningful testbed for both 
multilingual and low-resource evaluation. Our contributions are:

\begin{itemize}
    \item \transset{}, translated and localized prompts from the English 
    IFEval benchmark, carefully filtered and manually verified by native 
    speakers for cultural suitability and translation quality.
    
    \item \groundset{}, synthetically generated instructions grounded 
    in native Indic topics and content, manually verified by native 
    speakers. Unlike translated prompts, these reflect naturalistic 
    constraints with more real-world contexts.
    
    \item The first comprehensive multilingual evaluation of verifiable 
    instruction-following spanning diverse model families, parameter 
    sizes from 0.5B to 70B, and both reasoning and non-reasoning 
    configurations, and both open-weight and proprietary models.
    

    \item A comprehensive framework for extending verifiable instruction-following benchmarks to new languages with the availability of models tailored for translation or general use, and corpora of local contexts. 
\end{itemize}

Together, \transset{} and \groundset{} form around 800 human-verified examples 
 per language, providing complementary perspectives on multilingual 
instruction-following: the former enabling direct comparison against 
English baselines, the latter grounding evaluation in more natural constraints 
that translation alone cannot capture.

\section{Related Work}
Several recent works have extended the verifiable instruction-following paradigm to multilingual contexts. M-IFEval \cite{dussolle-etal-2025-ifeval} adapts the format to French, Japanese, and Spanish, highlighting that while models excel at formatting constraints, they struggle with script-specific tasks (e.g., ``do not use Katakana''). Similarly, CL-IFEval \cite{ojewale2025multi} expands this slightly to include French, Spanish, Hindi, Arabic, and Yoruba. These works mainly employ translations.

More comprehensive benchmarks have recently been proposed. MaXIFE \cite{liu-etal-2025-maxife}, introduces a dataset covering 23 languages with 795 prompts and $\sim$ 1.7k constraint templates, resulting in over 18,000 possible combinations. It features a diverse taxonomy derived from real examples, utilizing both LLM-as-a-judge and normal accuracy evaluation. However, one limitation is its reliance on translation without any localization. \citet{kamath2025benchmarking} craft new Hindi prompts with verifiable constraints through an in-house human annotation process guided by reference samples and Indian cultural themes. 

Marco-Bench-MIF \cite{zeng-etal-2025-marco} evaluates 30 languages with a specific focus on the distinction between translation and localization. They identify that machine-translated (MT) data underestimates model accuracy by 7–22\% compared to localized data, largely due to implicit English constraints (like capitalization) that become invalid in translation. While this offers deep insights into localization, it does not explore creating new prompts from grounded contexts. Further, it does not compare performance between reasoning and non-reasoning modes, leaving a gap in understanding how reasoning affects instruction adherence.

\section{Dataset Construction}
In this section we detail the process of obtaining the translations based on the English IFEval dataset \cite{ifeval} and LLM-based generations grounded in Indic contexts from \texttt{sangraha-verified} \cite{khan2024indicllmsuite}. An overview of the process is illustrated in Figure \ref{fig:overview_figure}.

\subsection{\transset{}}
\label{sec:translation}

We use Sarvam-Translate \cite{sarvam-translate} for the translations unless explicitly stated otherwise.

\paragraph{Preprocessing}
\label{sec:preprocessing}
First, we manually inspect the prompts in the IFEval dataset and assess their usability in non-English contexts. We localize the data with Indic-relevant themes wherever possible by simple modifications of named entities like `President of USA' to `Prime Minister of India' etc. For constraints involving explicit English responses (e.g., capitalization), we prepend the text ``Respond in English.'' to such prompts. 

Second, we spot cases that can have problematic translations -  abbreviations in the source prompt that may not make sense in non-English languages, improper translations due to translationese, translation collapse, tense-changes, and incomplete translations caused by the MT system. In most cases, we fix this by replacing problematic terms with another that would be easier to translate, however in few cases we chose to drop them. The exact details are outlined in Appendix \ref{sec:appendix:translation}.

In total, we arrive at 490 from the original 541 examples in this process. 

\paragraph{Keyword Extraction + Individual Keyword Translation}
We identify the prompts with keyword constraints and extract all the keywords. The keywords are first translated individually. The keywords that should not be translated (e.g, prompts where the response language is another language), are also first translated, but they are stored separately.

\paragraph{Pre-translation Insertion}
Then, we insert the keyword translation into the respective English prompt in place of the English keywords. We call this the `pre-translation'.

\paragraph{Full Translation}
Next, we translate the pre-translated prompts to the target language. Note that for non-keyword constraints, we translate them directly without modifications. We note that we preserve the non-English keywords as highlighted earlier. This is done by replacing the keyword automatically back to its original source after translation.

\paragraph{Automatic Verification}
Finally, using regular expressions, we check that the constraints are described properly in the translations. We manually fix these errors in formatting wherever necessary.

We discuss this in detail in Appendix \ref{sec:appendix:translation}. 

\subsection{\groundset{}}
\label{sec:grounding}
While translation provides a strong baseline for multilingual evaluation, it often lacks cultural grounding and may result in text that may be grammatically correct but constraints that are unnatural. 

We observe that this unnaturalness is especially true for keyword and length constraints. Required keywords are often arbitrary translations of English terms, which may not naturally occur in the target language's context. Further, certain keyword constraints (e.g., ``Start the response with X'') are often grammatically infeasible or highly unnatural in Indic languages due to their distinct word order. As for length constraints (e.g., ``Respond in X paragraphs'' or ``Respond in X sentences''), these may also not translate 1:1 to other languages.

To address this, we employ a context-grounded synthetic data generation pipeline. The process involves the following steps:
\begin{enumerate}
    \item \textbf{Define Constraint Types }
    We focus on a subset of six constraint types - Keyword Inclusion, Keyword Prohibition, Keyword Frequency, Paragraph Count, Sentence Count, and First Word, as these were identified to be the most problematic while translating from English. 

    \item \textbf{Context Mining } For valid keywords, we employ TFIDF-based sorting to extract top keywords that would mimic common, real-world keywords that people may use. Then, we retrieve text segments that match the defined constraint (e.g, contains five sentences, contains the keyword X two times etc.) by looping over all combinations (with an upper limit of 5 for frequency). 
    \item \textbf{Prompt Generation from Contexts} We use Gemini-2.5 \cite{gemini-2.5} to generate a user prompt based on the provided context that requires satisfying the given IFEval constraint, and an explicit mention of what the defined constraint is. We also include a randomly chosen instruction type from a list of tasks to diversify the generated prompts.    
\end{enumerate}


\subsection{Human Verification}
To ensure the validity and quality of the dataset, we implement a human-in-the-loop verification process for both \transset{} and \groundset{} subsets. We employ qualified, native speakers for each of the target languages as the annotators. For prompt verification, we assess the semantic validity of the prompt rather than strict grammatical perfection. Finally, we drop the prompts that are marked as incorrect - we do not perform any modifications to the prompts if they are incorrectly translated. 

We elaborate more on the annotation guidelines along with the final dataset statistics in the Appendix \ref{sec:appendix:verification}.

\section{Evaluation}
\begin{table*}[!t]
\resizebox{\linewidth}{!}{
\begin{tabular}{lcccccccccccccccc}
\toprule
Model & as & bn & gu & hi & kn & ml & mr & ne & or & pa & sa & ta & te & ur & en & Avg \\
\midrule
gemini/gemini-3-flash & \cellcolor{propbest}90.0 & 90.0 & 89.7 & 89.7 & 86.3 & 84.7 & 88.8 & 89.7 & 86.9 & 89.4 & 79.4 & 85.7 & 89.4 & 85.4 & \cellcolor{propbest}96.3 & 87.5 \\
gemini/gemini-3-pro & 89.4 & \cellcolor{propbest}92.5 & \cellcolor{propbest}91.3 & 91.0 & \cellcolor{propbest}91.3 & 86.9 & \cellcolor{propbest}91.3 & \cellcolor{propbest}92.2 & 86.0 & \cellcolor{propbest}93.2 & \cellcolor{propbest}86.3 & \cellcolor{propbest}89.7 & \cellcolor{propbest}90.3 & \cellcolor{propbest}89.1 & 95.3 & \cellcolor{propbest}90.0 \\
openai/gpt-5-2025-08-07 & 88.8 & 90.7 & 89.4 & \cellcolor{propbest}92.8 & 89.1 & \cellcolor{propbest}90.0 & 89.7 & 91.0 & \cellcolor{propbest}89.1 & 90.3 & 80.4 & 86.9 & 88.5 & 87.5 & 95.0 & 88.9 \\
openai/gpt-5-mini-2025-08-07 & 77.9 & 86.0 & 84.4 & 88.8 & 72.9 & 76.3 & 81.6 & 82.9 & 75.4 & 84.1 & 70.1 & 75.1 & 80.1 & 81.3 & 91.9 & 79.8 \\
\midrule
CohereLabs/aya-expanse-8b & 19.0 & 24.0 & 17.4 & 53.6 & 20.2 & 21.8 & 24.3 & 20.2 & 19.0 & 18.7 & 21.2 & 24.9 & 18.4 & 38.9 & 70.4 & 24.4 \\
CohereLabs/aya-expanse-32b & 33.0 & 42.7 & 33.0 & 70.4 & 30.8 & 39.9 & 32.7 & 37.7 & 35.5 & 32.1 & 29.9 & 41.4 & 26.5 & 57.6 & 77.3 & 38.8 \\
Qwen/Qwen3-0.6B & 23.7 & 30.2 & 23.7 & 30.5 & 28.7 & 23.4 & 27.7 & 25.2 & 22.7 & 26.5 & 24.0 & 21.5 & 25.9 & 28.0 & 65.4 & 25.8 \\
Qwen/Qwen3-1.7B & 32.4 & 43.3 & 35.5 & 48.3 & 38.3 & 35.5 & 39.2 & 37.7 & 41.1 & 34.0 & 28.7 & 31.8 & 34.9 & 44.9 & 75.4 & 37.5 \\
Qwen/Qwen3-4B & 54.2 & 62.6 & 54.2 & 66.4 & 54.2 & 48.0 & 57.9 & 58.9 & 47.4 & 46.7 & 48.3 & 48.6 & 53.3 & 59.8 & 85.0 & 54.3 \\
Qwen/Qwen3-8B & 59.5 & 67.3 & 63.9 & 74.1 & 67.9 & 61.7 & 66.7 & 64.8 & 52.6 & 57.6 & 54.5 & 60.8 & 62.0 & 64.8 & 88.2 & 62.7 \\
Qwen/Qwen3-14B & 69.2 & 72.6 & 72.9 & 78.8 & 75.1 & 69.5 & 70.4 & 72.3 & \cellcolor{openbest}74.5 & 70.1 & 64.5 & 65.4 & 69.8 & 70.4 & 87.8 & 71.1 \\
Qwen/Qwen3-32B & 71.7 & 73.8 & 74.8 & 78.5 & 74.1 & 73.5 & 75.7 & 73.5 & 70.4 & 67.6 & 64.8 & \cellcolor{openbest}73.2 & 72.0 & 75.4 & 87.2 & 72.8 \\
google/gemma-3-1b-it & 20.6 & 34.3 & 29.6 & 38.0 & 21.2 & 18.7 & 33.0 & 28.7 & 10.0 & 19.3 & 22.1 & 33.0 & 28.7 & 28.7 & 65.1 & 26.1 \\
google/gemma-3-4b-it & 50.8 & 73.8 & 65.4 & 75.1 & 53.9 & 56.7 & 68.2 & 65.4 & 38.0 & 50.2 & 44.5 & 67.6 & 64.8 & 67.3 & 81.9 & 60.1 \\
google/gemma-3-12b-it & 71.7 & 78.8 & 77.0 & 81.9 & 73.8 & 72.6 & 77.3 & 72.9 & 65.7 & 71.3 & 63.5 & 71.7 & 74.5 & \cellcolor{openbest}77.9 & 86.0 & 73.6 \\
google/gemma-3-27b-it & 72.3 & \cellcolor{openbest}81.0 & \cellcolor{openbest}80.4 & \cellcolor{openbest}85.0 & \cellcolor{openbest}76.3 & \cellcolor{openbest}77.3 & \cellcolor{openbest}77.6 & \cellcolor{openbest}79.4 & 69.2 & 75.7 & 64.8 & \cellcolor{openbest}73.2 & \cellcolor{openbest}76.0 & 77.6 & 88.2 & \cellcolor{openbest}76.1 \\
meta-llama/Llama-3.2-1B-Instruct & 16.5 & 17.8 & 13.4 & 27.1 & 15.9 & 14.0 & 22.1 & 22.4 & 18.1 & 16.5 & 20.6 & 15.9 & 15.6 & 20.6 & 54.2 & 18.3 \\
meta-llama/Llama-3.2-3B-Instruct & 23.4 & 34.9 & 22.7 & 42.7 & 23.7 & 17.8 & 30.8 & 29.0 & 17.1 & 25.9 & 18.4 & 22.1 & 28.4 & 39.2 & 77.6 & 26.9 \\
meta-llama/Llama-3.1-8B-Instruct & 38.9 & 44.9 & 29.0 & 53.6 & 39.2 & 31.8 & 41.7 & 45.5 & 28.0 & 35.8 & 32.4 & 32.1 & 33.0 & 49.2 & 81.6 & 38.2 \\
meta-llama/Llama-4-17B-Instruct & \cellcolor{openbest}73.2 & 76.0 & 77.0 & 77.3 & 75.1 & 73.5 & 73.8 & 75.4 & 66.4 & \cellcolor{openbest}76.0 & 51.7 & 69.5 & 72.0 & 76.6 & 89.4 & 72.4 \\
meta-llama/Llama-3.1-70B-Instruct & 60.4 & 67.0 & 55.5 & 71.0 & 59.5 & 55.5 & 60.8 & 64.8 & 52.6 & 59.8 & 51.1 & 54.5 & 61.4 & 74.1 & 88.5 & 60.6 \\
meta-llama/Llama-3.3-70B-Instruct & 72.0 & 77.0 & 73.2 & 79.4 & 72.0 & 66.7 & 74.8 & 74.8 & 63.5 & 70.1 & \cellcolor{openbest}67.3 & 69.2 & 72.3 & 76.3 & \cellcolor{openbest}91.6 & 72.0 \\
\midrule
Avg & 52.2 & 59.5 & 54.1 & 65.4 & 54.2 & 52.0 & 56.9 & 56.9 & 49.2 & 52.4 & 47.1 & 52.2 & 53.1 & 60.3 & 82.1 & 54.7 \\
\bottomrule
\end{tabular}
}
\caption{Overall Results for \transset{} (avg calculated only for Indic languages)}
\label{tab:main_results_translated}
\end{table*}

\subsection{Models}
We cover a diverse suite of models, spanning open-weights and proprietary architectures, across varying parameter sizes (from 1B to 70B+).

\begin{itemize}
    \item \textbf{Open-Weight Models:}
    \begin{itemize}
        \item \textit{Llama Family:} Llama-3.1 (8B, 70B), Llama-3.2 (3B), Llama-3.3 (70B), Llama-4 (17B-MoE-16E) \cite{llama-3, llama-4}.
        \item \textit{Gemma Family:} Gemma 3 (1B, 4B, 12B, 27B) \cite{gemini-3}.
        \item \textit{Aya Family:} Aya-Expanse (8B, 32B) \cite{aya-expanse}.
        \item \textit{Qwen Family:} Qwen 3 (0.6B, 1.7B, 4B, 8B, 14B, 32B) \cite{qwen-3}.
    \end{itemize}

    \item \textbf{Proprietary Models:}
    \begin{itemize}
        \item GPT-5, GPT-5-mini \cite{gpt-5}
        \item Gemini-3-Pro, Gemini-3-Flash \\\cite{gemini-3}
    \end{itemize}
\end{itemize}

\subsection{Metrics}

Following IFEval \cite{ifeval}, we follow the \texttt{prompt-level} and \texttt{instruction-level} accuracy for evaluation. In brief, the former calculates the percentage of prompts for which \textit{all} constraints are followed, and the latter calculates the percentage of prompts for which \textit{each individual instruction} is followed. We report the \texttt{loose} scores, which mitigate minor discrepancies in the response such as formatting.  

In our experiments, we observe that \texttt{instruction-level} accuracy scores are higher than \texttt{prompt-level}, however, the overall trends themselves are consistent, and hence, we focus primarily on prompt-level accuracy. In constraint-wise comparisons, accuracy is computed per instruction and then averaged across instructions, so the choice of reporting prompt-level versus instruction-level accuracy does not change the scores in these analyses.

\subsection{Evaluation}

As for the evaluation itself, the original IFEval scripts utilize NLTK \cite{bird-loper-2004-nltk} to verify structural constraints such as word, sentence, and paragraph counts. To reliably evaluate these constraints across our target languages, we modify this to integrate the Indic NLP Library \cite{kunchukuttan2020indicnlp} into our verification pipeline. This integration provides robust sentence segmentation and word tokenization optimized for Indic scripts to ensure that the subsequent scoring calculations remain precise.

To standardize the generation process across our diverse model suite, all evaluations are orchestrated using the Language Model Evaluation Harness \cite{eval-harness}. We enforce greedy decoding during all inference runs.

\section{Results and Analysis}
\subsection{\transset{}}

To analyze cross-lingual trends, we require a strictly parallel corpus. We therefore curate a common subset of prompts that are marked as correct by annotators across languages, amounting to 321 examples per language. The prompt-level loose accuracy scores are presented in Table \ref{tab:main_results_translated}. 

Across models, performance on English prompts (\texttt{en}) is consistently higher than the average over Indic languages (\texttt{Avg}), indicating a persistent cross-lingual gap in following verifiable instructions. This gap generally narrows for larger and more multilingual-oriented models, but remains pronounced for smaller and less multilingual models.

\paragraph{Effect of Model Parameters}

\begin{figure}[!h]
    \centering
    \includegraphics[width=\linewidth]{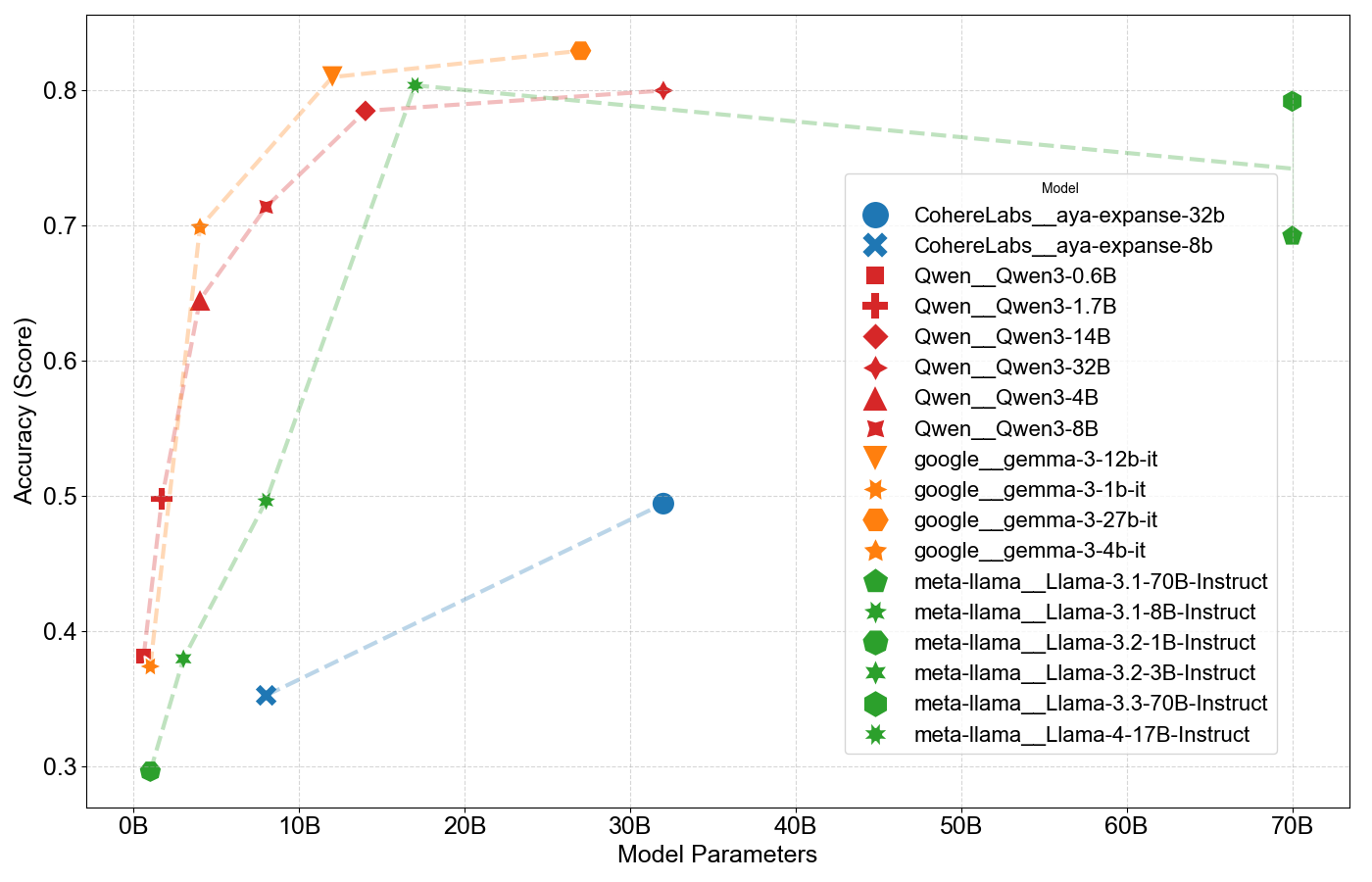}
    \caption{Effect of Increasing Model Parameters. The higher the better -- The Gemma-3-27B-IT model performs the best, followed by Llama-4-Scout-17B-16E. }
    \label{fig:model_parameters_effect}
\end{figure}

We find that increasing model parameters generally improves performance. Gemma-3-27B-IT wins overall followed by Gemma-3-12B-IT and Llama-4-Scout-17B-16E-Instruct. Qwen3-1.7B performs almost equally as the much larger Aya-Expanse-32B, possibly because of distilled thinking capabilities. At the smallest scale, Llama-3.2-1B-Instruct performs the poorest, and Gemma-3-1B-IT offers only marginal improvements, indicating that the \textbf{advantage of robust multilingual pre-training}, when comparing Gemma to Llama, \textbf{only materializes at larger scales} where the model has sufficient capacity to leverage it. Furthermore, performance plateaus around the 12B to 14B parameter range with limited gains thereafter (< 5), suggesting verifiable instruction adherence requires a minimum baseline of parameter capacity.\\
\vspace{-15pt}
\begin{figure}[!h]
\begin{tcolorbox}[colback=openbest, colframe=openbest, fontupper=\small, arc=4pt, boxrule=0pt]
Note: We observe that the Indic performance is generally lower than the English baseline. Hence, to quantify how small or large this gap is to achieve parity, we present the subsequent analyses by viewing the disparity between the Indic language score and the English score(`$\Delta$'; ranging -1 to 1).

The lower the \underline{absolute} gap, the better.
\end{tcolorbox}
\end{figure}
\vspace{-15pt}

\paragraph{Model Family Performance}

\begin{figure}[!h]
    \centering
    \includegraphics[width=\linewidth]{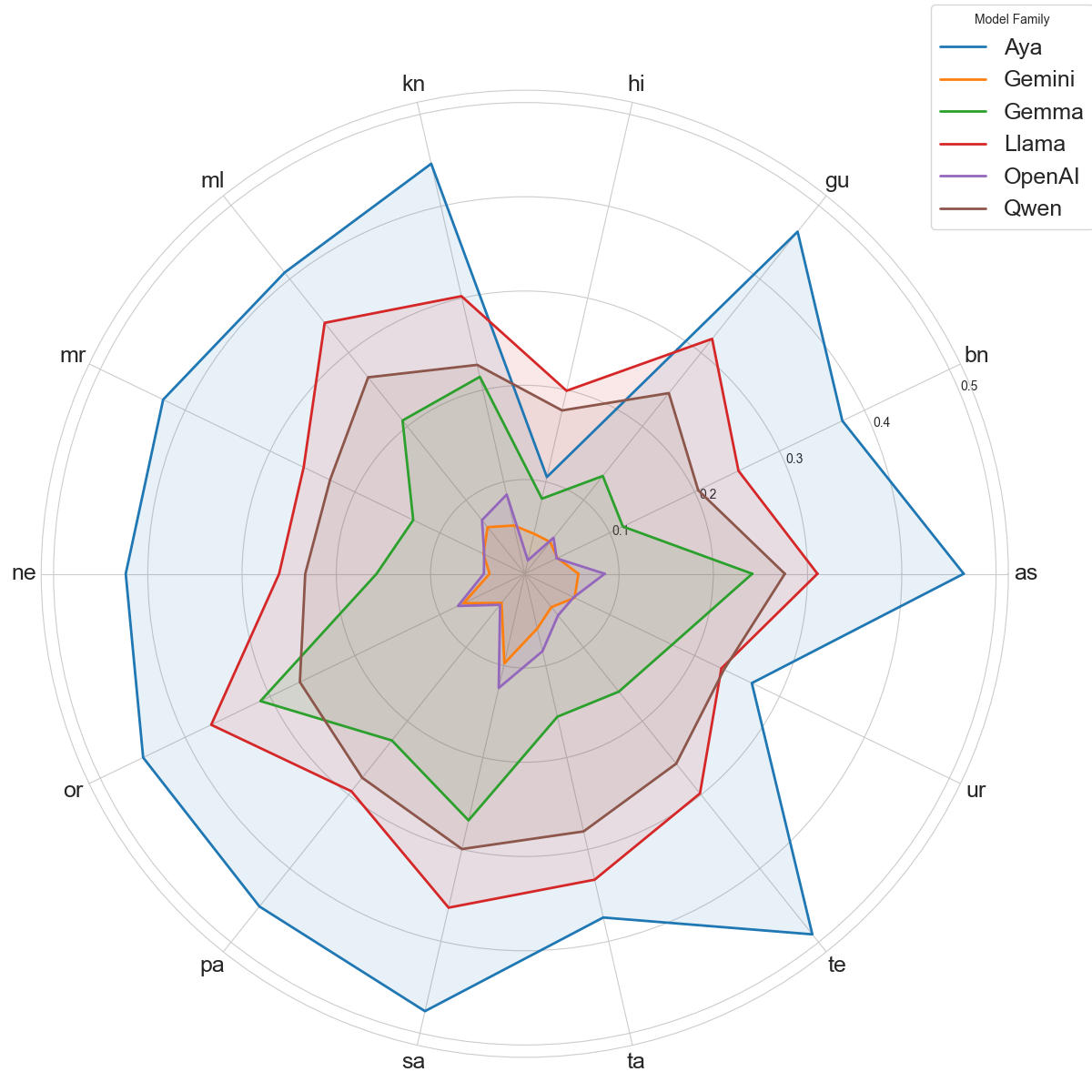}
    \caption{Indic - English disparity comparing Language vs Model Family. The lower the better -- the Aya family exhibits the highest gap in general, but performs moderately well for Hindi, Bengali, Urdu, and Tamil.}
    \label{fig:indic_vs_english_heatmap}
\end{figure}

The \textbf{Gemma family} consistently exhibits the \textbf{smallest performance gap with English}  across all languages (e.g., $\Delta<0.15$ for Hindi and Bengali), suggesting its training objective is the most robust to cross-lingual inputs (especially given that its largest variant, i.e. 27B is less than half the size of the largest evaluated models, i.e. 70B).

The Qwen and Llama families occupy a competitive middle tier, with Qwen frequently outperforming Llama by maintaining slightly lower disparities in languages like Bengali and Gujarati. Conversely, the \textbf{Aya family} displays the highest degradation, with gap scores exceeding $0.40$ for most low-resource languages, indicating a \textbf{significant struggle to transfer instruction-following precision despite its multilingual focus}. Notably, Hindi remains the most robust language across all model families.

\paragraph{Individual Constraint Capabilities}

\begin{figure}[!h]
    \centering
    \includegraphics[width=\linewidth]{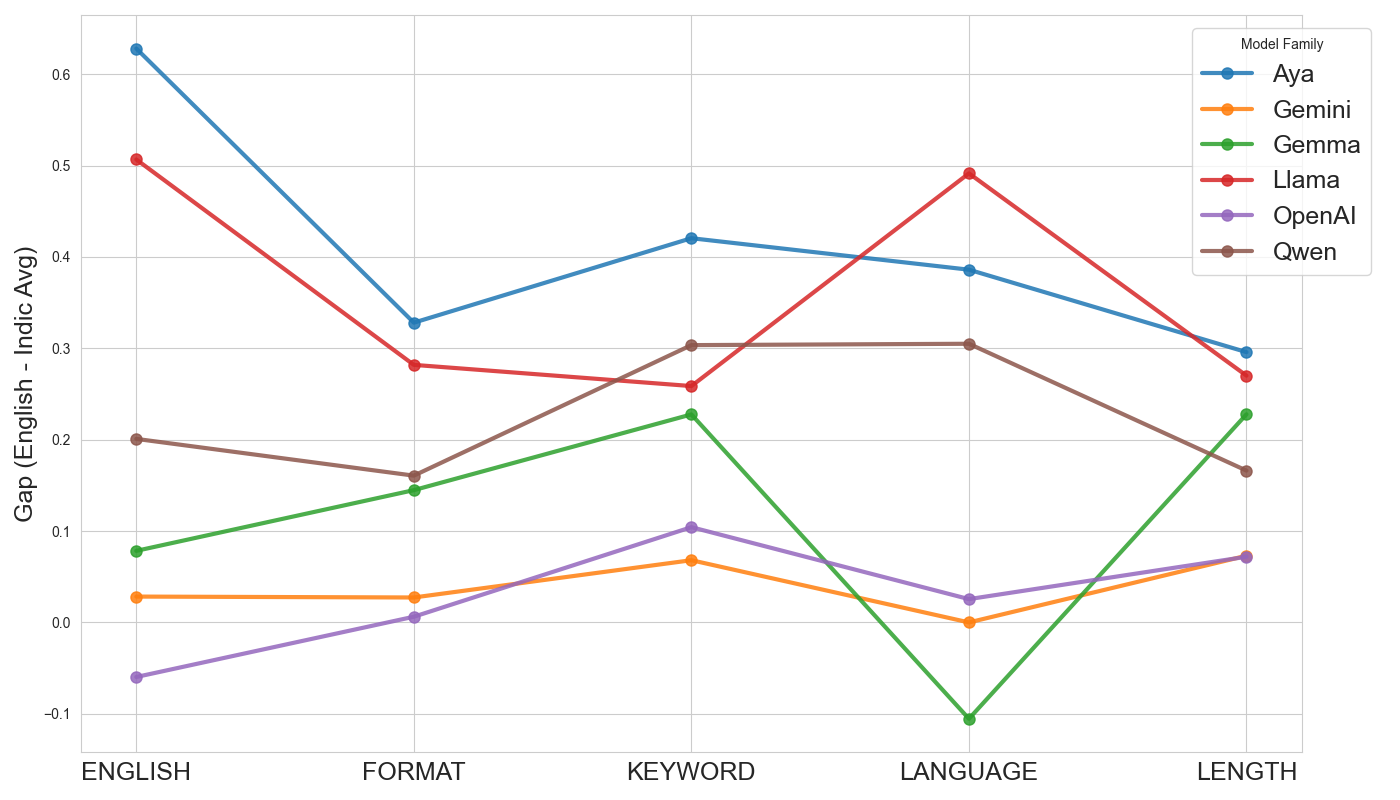}
    \caption{Indic - English gap comparing Model Family vs Instruction Category. Lower the better -- the Llama family and Gemma family perform the best and worst respectively on \textsc{Language} constraint within the family.}
    \label{fig:family_vs_category}
\end{figure}

To provide a more granular understanding of model capabilities across different constraints, we examine the average performance for each constraint type. The standard IFEval framework consists of 25 distinct instruction constraints. Analyzing these constraints individually would be difficult and may yield fragmented insights. Hence, we aggregate these 25 constraints into five high-level semantic categories: \textsc{English}, \textsc{Keyword}, \textsc{Format}, \textsc{Length}, and \textsc{Language} to find broader patterns. The specific mapping of individual IFEval instructions to these five categories is detailed in Table \ref{tab:constraint_categories}.

\begin{table*}[!t]
\centering
\resizebox{\linewidth}{!}{
\begin{tabular}{lccccccccccccccc}
\toprule
Model & as & bn & gu & hi & kn & ml & mr & ne & or & pa & sa & ta & te & ur & Avg \\
\midrule
gemini/gemini-3-flash & 64.6 & 67.0 & 60.7 & 61.9 & 58.1 & 56.2 & 62.9 & 66.9 & 62.8 & 64.5 & 55.6 & 64.9 & 64.3 & 59.6 & 62.1 \\
gemini/gemini-3-pro & 62.3 & 66.0 & 61.0 & 58.9 & 61.0 & 59.4 & 62.0 & 64.2 & 67.9 & 65.5 & 50.9 & 66.0 & 70.4 & 60.2 & 62.6 \\
openai/gpt-5-2025-08-07 & 73.2 & \cellcolor{propbest}72.6 & \cellcolor{propbest}77.5 & \cellcolor{propbest}77.9 & \cellcolor{propbest}73.6 & \cellcolor{propbest}70.6 & 78.0 & 72.4 & \cellcolor{propbest}80.4 & \cellcolor{propbest}78.5 & 78.3 & 78.1 & 78.4 & 72.7 & \cellcolor{propbest}75.9 \\
openai/gpt-5-mini-2025-08-07 & \cellcolor{propbest}75.5 & 71.2 & 74.5 & 75.5 & 71.5 & 68.2 & \cellcolor{propbest}81.7 & \cellcolor{propbest}75.7 & 78.8 & 77.0 & \cellcolor{propbest}79.5 & \cellcolor{propbest}79.1 & \cellcolor{propbest}78.7 & \cellcolor{propbest}73.6 & 75.8 \\
\midrule
CohereLabs/aya-expanse-8b & 38.1 & 46.0 & 38.2 & 48.4 & 41.1 & 43.2 & 42.0 & 40.5 & 39.1 & 43.8 & 25.6 & 52.1 & 39.5 & 40.4 & 41.3 \\
CohereLabs/aya-expanse-32b & 59.9 & 56.2 & 46.1 & 54.9 & 45.8 & 55.5 & 53.1 & 50.7 & 47.8 & 42.5 & 36.0 & 58.4 & 48.0 & 42.4 & 49.8 \\
Qwen/Qwen3-0.6B & 48.6 & 53.5 & 50.7 & 56.9 & 47.5 & 42.7 & 56.0 & 54.8 & 32.3 & 38.6 & 43.5 & 49.3 & 51.2 & 47.8 & 48.1 \\
Qwen/Qwen3-1.7B & 59.9 & 56.7 & 51.5 & 58.4 & 48.1 & 45.6 & 62.6 & 61.9 & 44.8 & 37.9 & 53.9 & 55.1 & 49.9 & 54.0 & 52.9 \\
Qwen/Qwen3-4B & 71.2 & 69.4 & 61.3 & 63.7 & 58.1 & 55.2 & 62.6 & 62.8 & 62.2 & 48.7 & \cellcolor{openbest}67.0 & 61.2 & 64.8 & 59.9 & 62.0 \\
Qwen/Qwen3-8B & 73.2 & 70.2 & 65.2 & 64.3 & 67.5 & 60.9 & 65.4 & 73.0 & 68.2 & 63.3 & 62.8 & 71.2 & 67.2 & 60.2 & 66.6 \\
Qwen/Qwen3-14B & \cellcolor{openbest}74.3 & \cellcolor{openbest}72.1 & \cellcolor{openbest}71.3 & \cellcolor{openbest}73.1 & \cellcolor{openbest}69.2 & \cellcolor{openbest}70.3 & 69.7 & \cellcolor{openbest}75.7 & 67.4 & \cellcolor{openbest}67.5 & 64.3 & \cellcolor{openbest}76.7 & 68.5 & 60.2 & \cellcolor{openbest}70.0 \\
Qwen/Qwen3-32B & 73.2 & 70.2 & 64.2 & 62.4 & 61.3 & 61.5 & 66.6 & 70.4 & \cellcolor{openbest}69.3 & 64.8 & 61.3 & 70.0 & \cellcolor{openbest}71.2 & 57.6 & 66.0 \\
google/gemma-3-1b-it & 45.9 & 54.8 & 58.6 & 58.2 & 41.9 & 43.8 & 55.7 & 56.0 & 20.6 & 43.0 & 43.5 & 60.5 & 55.2 & 49.9 & 49.1 \\
google/gemma-3-4b-it & 61.5 & 57.2 & 60.7 & 64.1 & 56.0 & 57.3 & 64.0 & 63.3 & 57.6 & 59.7 & 65.8 & 60.5 & 61.6 & 54.0 & 60.2 \\
google/gemma-3-12b-it & 58.4 & 63.3 & 63.4 & 63.5 & 63.0 & 60.7 & 61.7 & 66.9 & 62.0 & 62.1 & 54.2 & 67.9 & 68.0 & 53.7 & 62.0 \\
google/gemma-3-27b-it & 69.3 & 63.6 & 64.7 & 60.0 & 59.2 & 63.0 & 66.0 & 65.1 & \cellcolor{openbest}69.3 & 65.8 & 59.2 & 60.7 & 66.4 & 54.3 & 63.3 \\
meta-llama/Llama-3.2-1B-Instruct & 42.4 & 63.6 & 65.0 & 60.8 & 42.2 & 47.1 & 53.7 & 57.8 & 31.5 & 46.7 & 50.9 & 55.1 & 61.6 & 59.4 & 52.7 \\
meta-llama/Llama-3.2-3B-Instruct & 58.0 & 71.4 & 53.6 & 64.8 & 64.8 & 56.8 & \cellcolor{openbest}70.9 & 63.0 & 48.4 & 59.7 & 44.6 & 71.6 & 70.1 & \cellcolor{openbest}61.7 & 61.4 \\
meta-llama/Llama-3.1-8B-Instruct & 62.6 & 68.7 & 63.4 & 65.2 & 62.8 & 62.8 & 61.1 & 71.3 & 56.2 & 62.6 & 65.2 & 68.8 & 70.7 & 56.4 & 64.1 \\
\midrule
Avg & 61.7 & 63.9 & 60.6 & 62.8 & 57.5 & 56.9 & 62.9 & 63.8 & 56.1 & 57.5 & 55.9 & 64.6 & 63.5 & 56.7 & 60.3 \\
\bottomrule
\end{tabular}
}
\caption{Overall Results for \groundset{}}
\label{tab:main_results_grounded}
\end{table*}

We observe that models consistently demonstrate \textbf{strong adherence to \textsc{Format} constraints}. The Gemma and Qwen families achieve the highest performance in this category, followed closely by the Llama family. The Aya model family exhibits significantly lower performance across all categories compared to other evaluated families (see Figure \ref{fig:family_vs_category}).

\begin{figure}[!h]
    \centering
    \includegraphics[width=\linewidth]{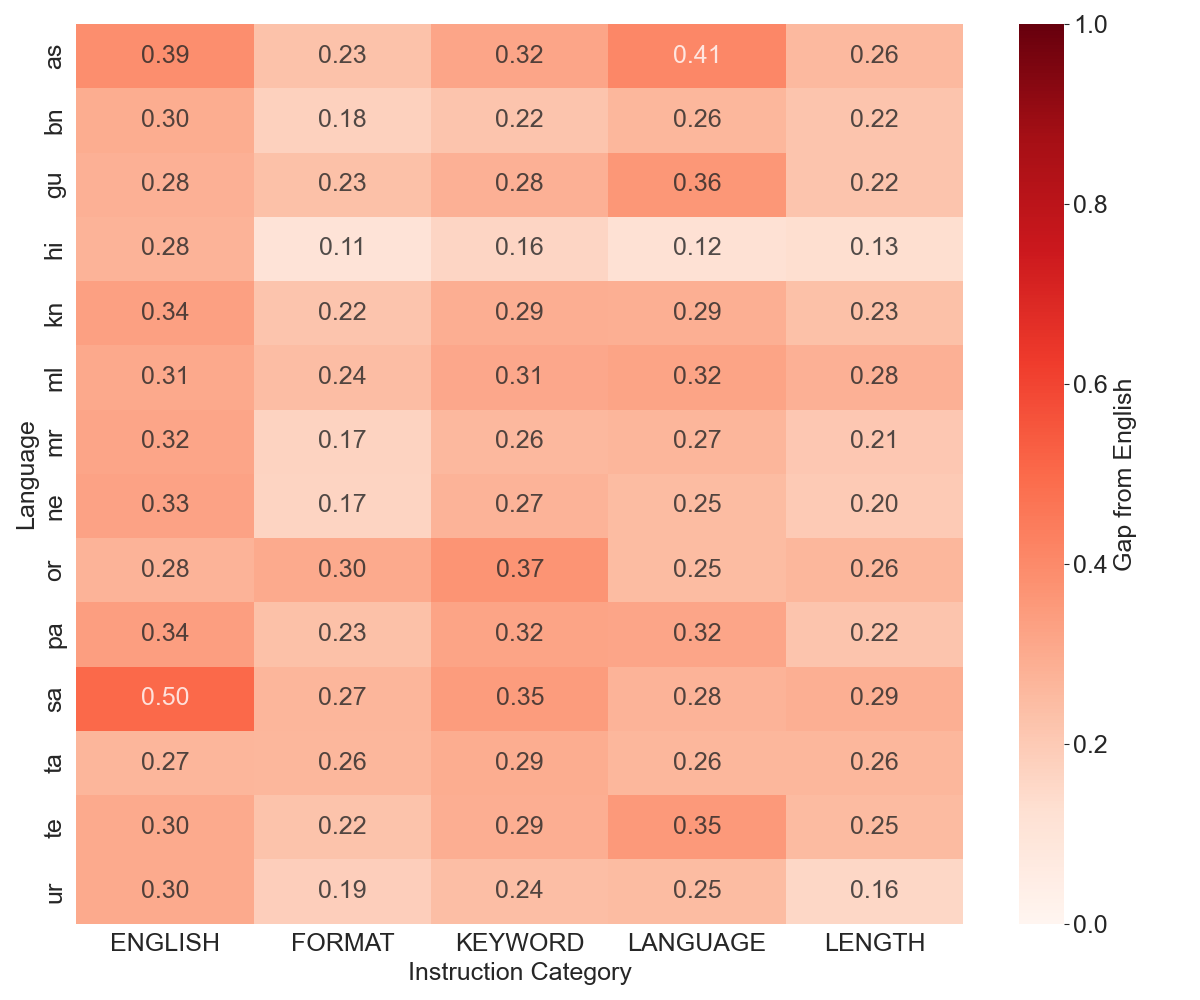}
    \caption{Indic - English disparity comparing Instruction Category vs Language. Lower the better -- Hindi (`hi') is visually the lightest (lowest $\Delta$) across all categories horizontally and darkest (highest $\Delta$) across all languages vertically. }
    \label{fig:language_vs_category}
\end{figure}

Moving to individual languages (refer Figure \ref{fig:language_vs_category}), we find that among all evaluated languages, Hindi exhibits the minimum $\Delta$ across the \textsc{Format}, \textsc{Language}, and \textsc{Length} constraints. This indicates that high-resource Indic languages like Hindi are nearing parity with English in structural instruction following. The most significant performance degradation relative to English is observed when models are prompted in Sanskrit, particularly when the constraints require an English response. 

\paragraph{Keyword Inclusion and Exclusion}

Within the \textsc{Keyword} category, we observe a strong performance disparity between negative constraints (Exclusion) and positive constraints (Inclusion). Across all model families, models demonstrate high proficiency in avoiding specific keywords when instructed. Conversely, models significantly struggle with including keywords, be it at least once or a certain number of times. 

This gap suggests that while \textbf{models possess the semantic understanding to suppress concepts} (negative constraints), they \textbf{face challenges in generating the precise lexical surface forms required for inclusion} (positive constraints).

\paragraph{Cross-lingual Instruction Following}
Given that models demonstrate high proficiency in English generation when prompted in English, we analyze their performance when the same English output is requested via Indic prompts (i.e., comparing the \textsc{English} column in Figure \ref{fig:family_vs_category} and Figure \ref{fig:language_vs_category} to other columns). Interestingly, we find that \textbf{shifting the prompt language introduces a performance gap, despite the general higher English capabilities}. Even high-resource languages like Hindi exhibit their maximum performance gap in this category ($\Delta=0.28$) overall, significantly higher than all other constraints where $\Delta=0.11-0.16$ (Figure \ref{fig:language_vs_category}). 
The Llama and Aya families suffer massive degradation ($\Delta > 0.50$). Upon manual inspection of the generations, we find that \textbf{models fail in such cross-lingual English instruction following} either when the \textbf{model continues to respond in the prompt language (Indic)}, thereby failing the English response constraint, or \textbf{respond in English but do not follow the capitalization constraints}. 

The Gemma family effectively closes this gap having $\Delta=0.08$ (Figure \ref{fig:family_vs_category}), demonstrating remarkable multilingual understanding and generation. To our surprise, we find that the Gemma family performs worse on the \textsc{English} constraint than on the \textsc{Language} constraint, with the \textsc{English} category $\Delta = -0.11$, the only case it happens to be negative. This negative difference signifies that \textit{prompting in English} and instructing to respond in another language is \textit{worse} than \textit{prompting in the target language} and instructing to respond in another language. However, a qualitative analysis reveals that \textbf{Gemma models frequently provide translations or transliterations in parentheses for readability} \textit{when the prompt is in English}. While helpful for human readers, this behavior \textbf{inadvertently violates strict language constraints}, thereby reducing the quantitative performance.

These discrepancies suggests that for many model families, cross-lingual instruction-following capabilities lag significantly behind their primary in-language capabilities, revealing a brittleness in how they ground Indic instructions to English outputs.

\paragraph{Open-Weight vs Proprietary}

\begin{figure}[!h]
    \centering
    \includegraphics[width=\linewidth]{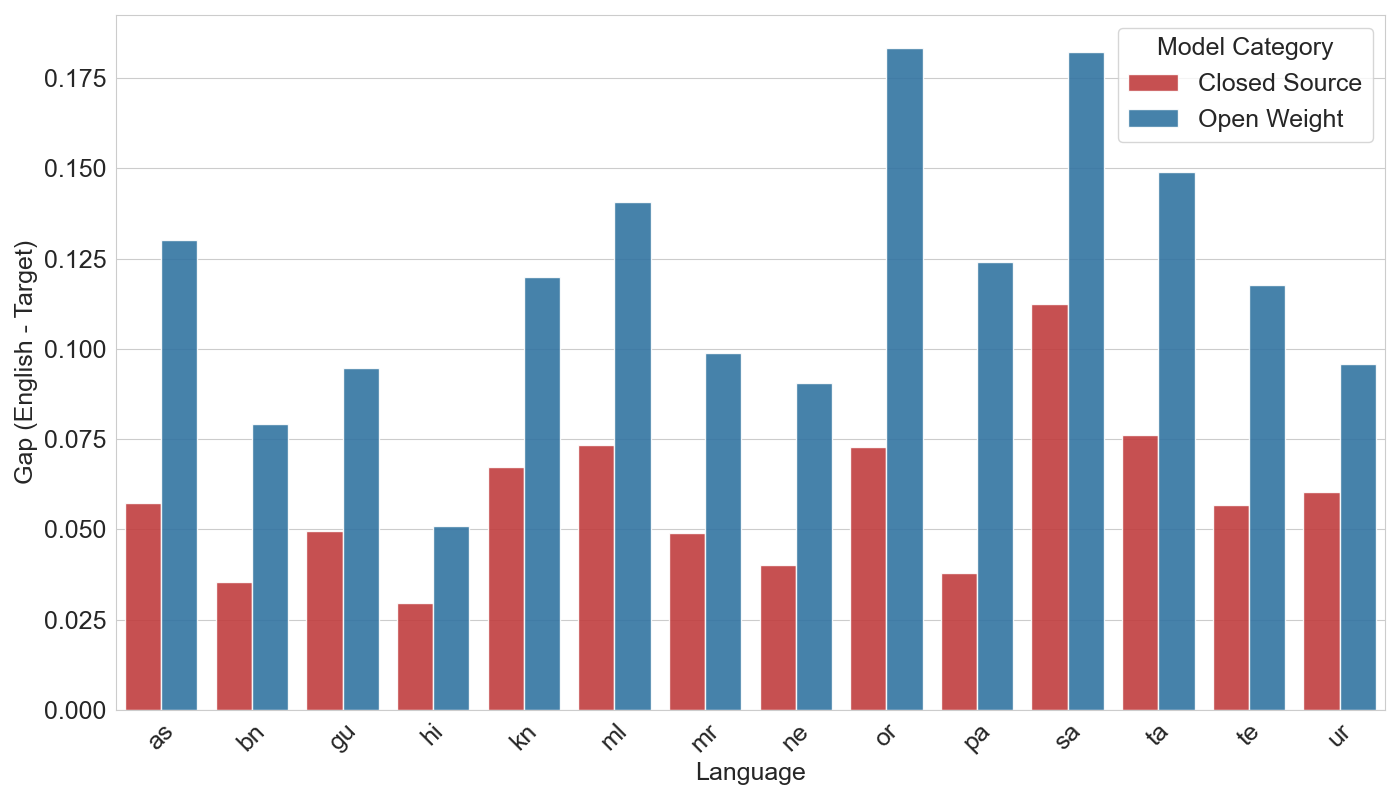}
    \caption{Indic vs English Disparity comparing Open-Weight and Proprietary. Lower the better -- Hindi (`hi') exhibits the lowest gap with English in both proprietary (red) and open-weight settings (blue).}
    \label{fig:open_and_closed_source}
\end{figure}

For a better comparison of open-weight and proprietary models, we take the top two performing models from open-weight (i.e., Gemma-27B-IT and Llama-3.3-70B-Instruct) so as to not skew the results. We view the English vs Indic disparity (i.e., $\Delta$) averaging across the two models selected for open-weight and four models in proprietary. We find that proprietary models consistently maintain a significantly lower $\Delta$ across all evaluated languages. While \textbf{open-weight models exhibit substantial variance depending on the language resource availability} (peaking at a disparity of $\Delta \approx 0.18$ for languages like Odia and Sanskrit), \textbf{their ability to close the gap is comparable to proprietary models in high-resource scenarios}. Notably, for Hindi, the open-weight $\Delta$ narrows to its minimum ($\Delta \approx 0.03$), nearly converging with the proprietary baseline. 

However, we note that when viewing the absolute accuracy scores (see Table \ref{tab:main_results_translated}), \textbf{there is still a 10+ score margin between the open-weight and proprietary models}, for verifiable constraints. Larger parameter sizes could potentially close this absolute performance gap, however, this is currently not visible, given that the best performing open-weight model is Gemma at only 27B.
\paragraph{Thinking vs Non-Thinking Modes}

\begin{figure}[!h]
    \centering
    \includegraphics[width=\linewidth]{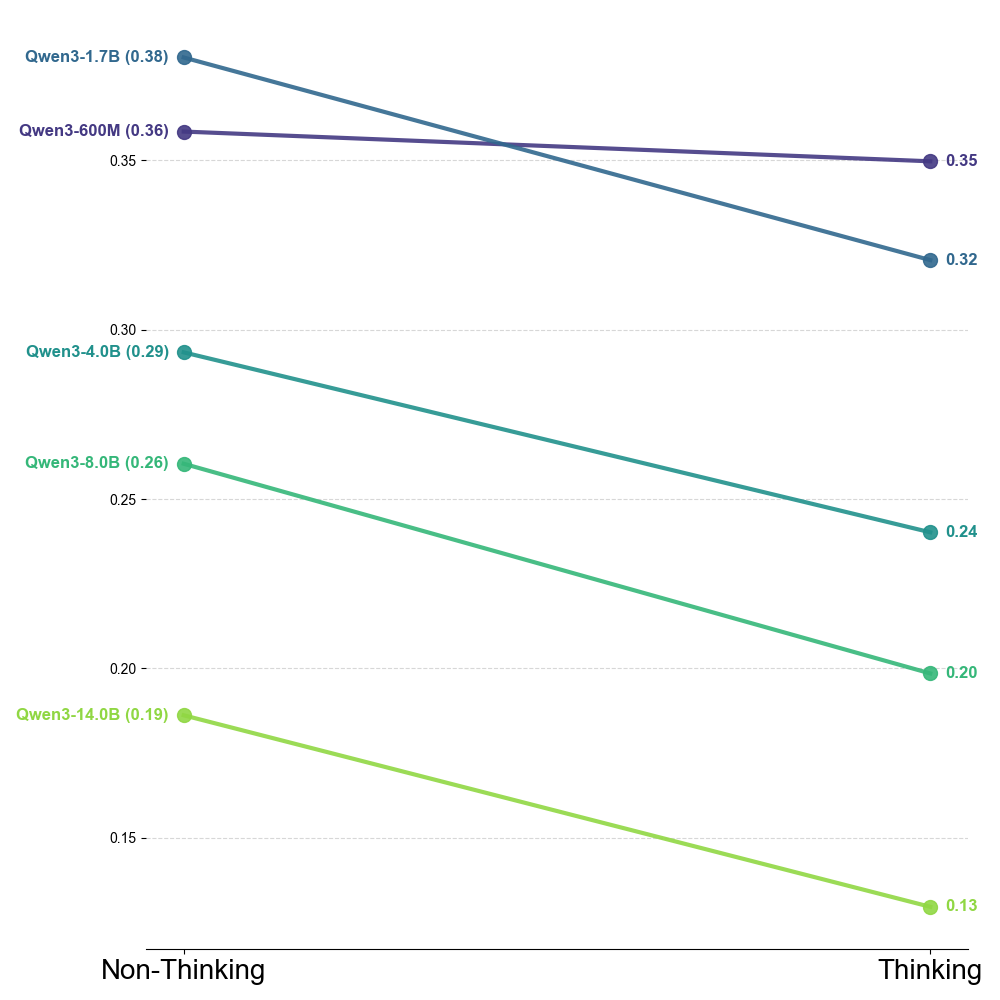}
    \caption{Indic vs English Disparity comparing Thinking and Non-Thinking Modes. A steeper negative slope indicates that thinking more effectively narrows the performance gap between English and Indic languages.}
    \label{fig:thinking_and_non_thinking}
\end{figure}

We choose the Qwen3 models as they support both thinking and non-thinking modes within the same architecture, allowing for a controlled ablation for varying parameters (0.6-32B). As illustrated in Figure \ref{fig:thinking_and_non_thinking}, \textbf{enabling the ``Thinking'' mode consistently narrows the performance disparity between English and Indic languages} across nearly all model scales, indicating that the additional thinking cost allows the model to better bridge the gap. At the lowest scale, the 600M model exhibits a negligible improvement (an almost flat slope from 0.36 to 0.35), suggesting that sub-billion parameter models may lack the capacity to effectively leverage thinking or cross-lingual transfer. 

Further, the roughly parallel trajectories (similar slope) of the thinking and non-thinking configurations demonstrate that the \textbf{performance benefit derived from the thinking mechanism remains constant and independent of the overall parameter count}. This is true when viewing the absolute scores too ($\sim$ 5-8 accuracy improvement when enabling thinking).

\subsection{\groundset{}}
\transset{}, being parallel, allows for a direct comparison across languages. However, it relies on the assumption that translated prompts preserve the naturalness of the original instruction and previous research show that translations can be biased towards the original English-contexts \cite{zeng-etal-2025-marco}. We report the evaluation results (loose prompt-following acc.) in Table \ref{tab:main_results_grounded}. 

It is important to note that the results presented in this section are not directly comparable to the \transset{} scores discussed previously. The prompts in \groundset{} differ significantly in complexity, topic, and structure. Consequently, there is no direct English baseline for these specific prompts; We focus exclusively on the absolute performance of models.

Further, the scores should not be viewed as direct comparison among Indic languages themselves as the prompts are entirely different for each language. This prompt diversity is necessary to capture the structural and cultural variance of Indic languages, which span multiple language families, and utilize different word orders.




\section{Conclusion}
We introduce \mainset{} to benchmark verifiable instructions for 14 Indic languages. 
By pairing two complementary datasets with parallel prompts in \transset{} and natural constraints in \groundset{}, this benchmark provides a comprehensive framework for assessing instruction following in multilingual and low-resource settings.
Our evaluations reveal a persistent performance gap between English and Indic languages even for simple, rule-based instructions. High-resource languages like Hindi consistently exhibit a narrower gap and approach English parity. Among open-weight architectures, the Gemma family demonstrates the highest cross-lingual robustness, whereas proprietary models maintain a distinct overall advantage across the broader language spectrum. Models generally adhere to structural formatting instructions but face substantial difficulties with precise lexical generation, such as keyword inclusion, and cross-lingual tasks. Increasing model capacity and employing thinking modes consistently reduce the performance gap between English and Indic languages. 

\section{Limitations}
Our grounded benchmark, \groundset{}, explores only a subset of possible instruction constraint types and is therefore not exhaustive. In particular, we restrict our study to the 25 constraint categories defined in the original IFEval benchmark, whereas future work could incorporate more fine-grained and linguistically motivated constraints, such as romanization requirements, explicit specification of numeral formats (e.g., digits versus words), and system-of-units constraints (e.g., Indian vs International numeral systems, Imperial vs Metric system). In addition, our annotations were produced by a single annotator due to resource constraints; incorporating multiple annotators in future work would enable stronger verification through inter-annotator agreement and improve annotation reliability. Our exploration of reasoning capabilities is restricted to the Qwen models, leaving the evaluation of other models with thinking mechanisms for future work. 

\bibliography{custom.bib}

\clearpage

\appendix
\section{Dataset Details}
The following is the description of the columns in the original IFEval benchmark \cite{ifeval}: 
\begin{enumerate}
    \item \textbf{key:} Unique integer identifier for each evaluation instance.
    \item \textbf{prompt:} The natural language instruction presented to the model.
    \item \textbf{instruction\_id\_list} List of instruction identifiers that specify which format or constraint checks apply to the prompt.
    \item \textbf{kwargs}: A list of parameter values associated with each constraint (e.g., required word, sentence count) in instruction\_id\_list.
\end{enumerate}

We implement the same format for \mainset{} to allow a streamlined and compatible evaluation framework across all the languages. 

We show some examples of the curated data in Table \ref{tab:translated_examples} and Table \ref{tab:grounded_examples} for \transset{} and \groundset{} respectively.

Table \ref{tab:constraint_categories} presents the categorization of constraints used in our analyses. We note that this is different from the Tier 1 categorization in IFEval.

\begin{table}[h]
\centering
\renewcommand{\arraystretch}{1.2}
\resizebox{\linewidth}{!}{
\begin{tabular}{p{2cm} p{11cm}}
\toprule
\textbf{Category} & \textbf{Instruction Constraints} \\
\midrule
\textsc{English} & change\_case:capital\_word\_frequency, change\_case:english\_capital, change\_case:english\_lowercase \\
\midrule
\textsc{Keyword} & combination:repeat\_prompt, detectable\_content:postscript, detectable\_format:constrained\_response, detectable\_format:multiple\_sections, detectable\_format:title, keywords:existence, keywords:forbidden\_words, keywords:frequency, length\_constraints:nth\_paragraph\_first\_word, punctuation:no\_comma, startend:end\_checker, keywords:letter\_frequency, startend:quotation \\
\midrule
\textsc{Format} & combination:two\_responses, detectable\_content:number\_placeholders, detectable\_format:json\_format, detectable\_format:number\_bullet\_lists, detectable\_format:number\_highlighted\_sections \\
\midrule
\textsc{Length} & length\_constraints:number\_paragraphs, length\_constraints:number\_sentences, length\_constraints:number\_words \\
\midrule
\textsc{Language} & language:response\_language \\
\bottomrule
\end{tabular}
}
\caption{Taxonomy of IFEval instruction constraints grouped into five high-level categories for analysis.}
\label{tab:constraint_categories}
\end{table}

\begin{table*}[t]
    \centering
    \includegraphics[width=\linewidth]{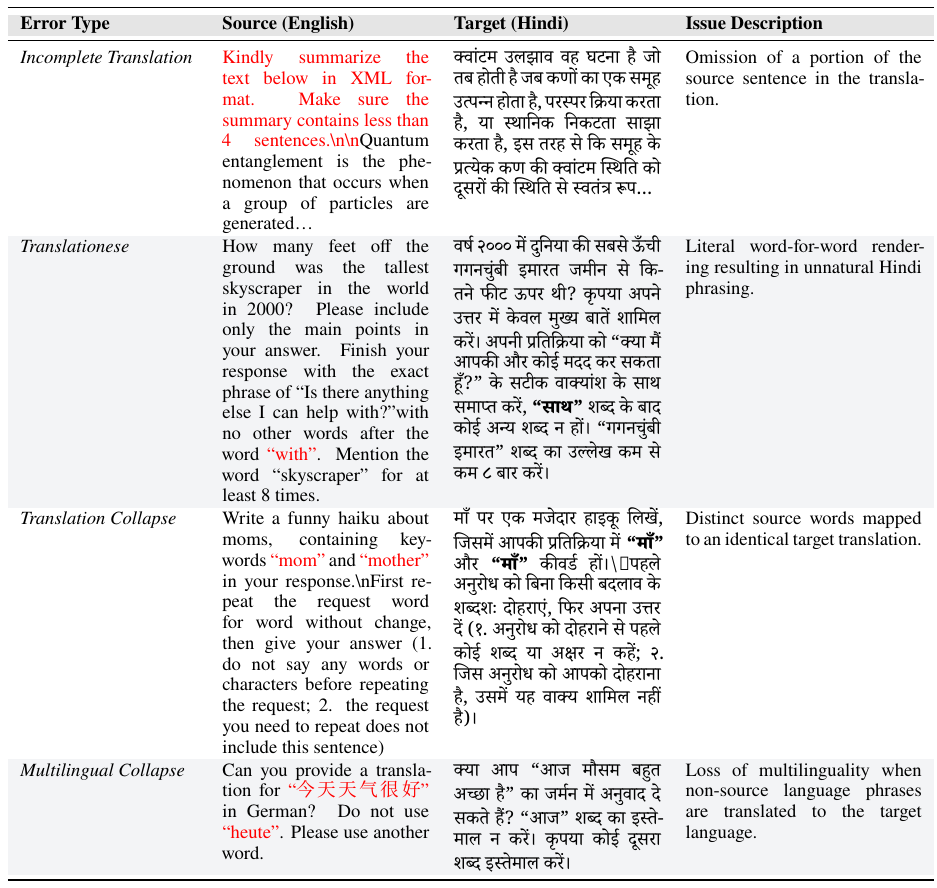}
    \caption{Illustrative examples of translation errors observed during our automatic translation from the English prompts.}
    \label{tab:translation_errors}
\end{table*}

\section{Translation}
\label{sec:appendix:translation}
We describe the translation process outlined in Section \ref{sec:translation} in more detail here.

Once the dataset is pre-processed, we cannot directly translate the prompts as-is, especially for keyword-constraints, since we need to ensure that the keyword in consideration (which is present in the metadata) is the same as the one used in the prompt. Translating the prompt and keyword separately may result in different words for the same keyword (because context is lost), and extracting the keyword after translation is not so straightforward (because there is no explicit marker and using LLMs may be unreliable for low-resource languages).

Further, if there is any non-English language content in the prompt, these should not be translated -- we only want to translate the keywords for those prompts whose response are expected to be in English. For example, if the response language is constrained to be German for a certain prompt, and the user also requests not to include the word ``Flughafen'', we cannot translate that word to our target language, otherwise this would render the German response constraint useless. Thus, we mark such rows for keyword retention during translation.

We find that the dataset being English-centric, does not contain direct metadata that indicate that the response language in English (often found in capitalization-based constraints). This is so because irrespective of whether we constrain the LLM to respond in English, the response will always be in English, since our instructions are in English. In general, LLMs are implicitly constrained to the language of the prompt.
Translating the prompts involving capitalization constraints needs to preserve this implicit English response constraint in the prompt. To address this, we prepend the text ``Respond in English.'' to such prompts before translating them.


In most cases, we modify the prompt such that these errors are minimized during automatic translation. 

For instance, incomplete translations often arise when the prompt includes task signals such as ``summarize'', ``translate'' etc. In such cases, the translation model executes the implied task (e.g., summarization) in the target language itself. This is more prevalent when `\textbackslash{n}' is present in the prompt. Furthermore, multilingual prompts also suffer from a similar failure mode, where non-source language (i.e., non-English) phrases get translated into the target language, instead of a faithful translation of the full prompt which preserves the respective languages. We note down the key of such commonly mistranslated prompts, and apply an additional pre-processing in the form of a simple text manipulation (e.g., remove `\textbackslash{n}' or the problematic keyword before translation and place again after translation).

For translation collapse, ie. when two keywords get translated into the same keyword (see Table \ref{tab:translation_errors}, both ``mother'' and ``mom'' get mapped to ``\foreignlanguage{hindi}{माँ}'', we collect the list of keywords that are susceptible to this failure through a simple script comparing source keywords and keywords after translation in the same prompt, and instruct Gemini-2.5-Flash to provide alternative words for those keywords that are semantically equivalent and keep those instead. As for examples similar to the one provided in the translationese error, we chose to drop such examples as post-processing them would make the process more complicated and such examples were very few (<5).

Finally, the resulting dataset comprises 490 examples from the original 541 examples.

\begin{figure*}[h]
    \centering
    \includegraphics[width=1.0\linewidth]{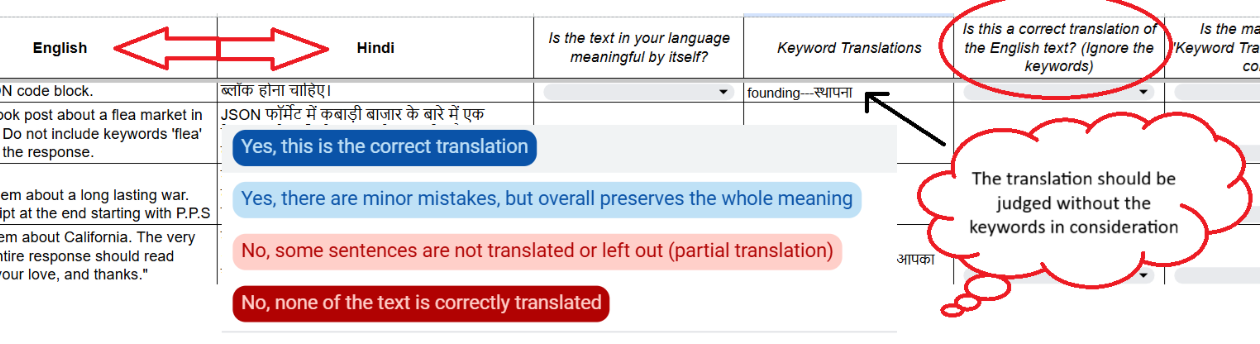}
    \includegraphics[width=1.0\linewidth]{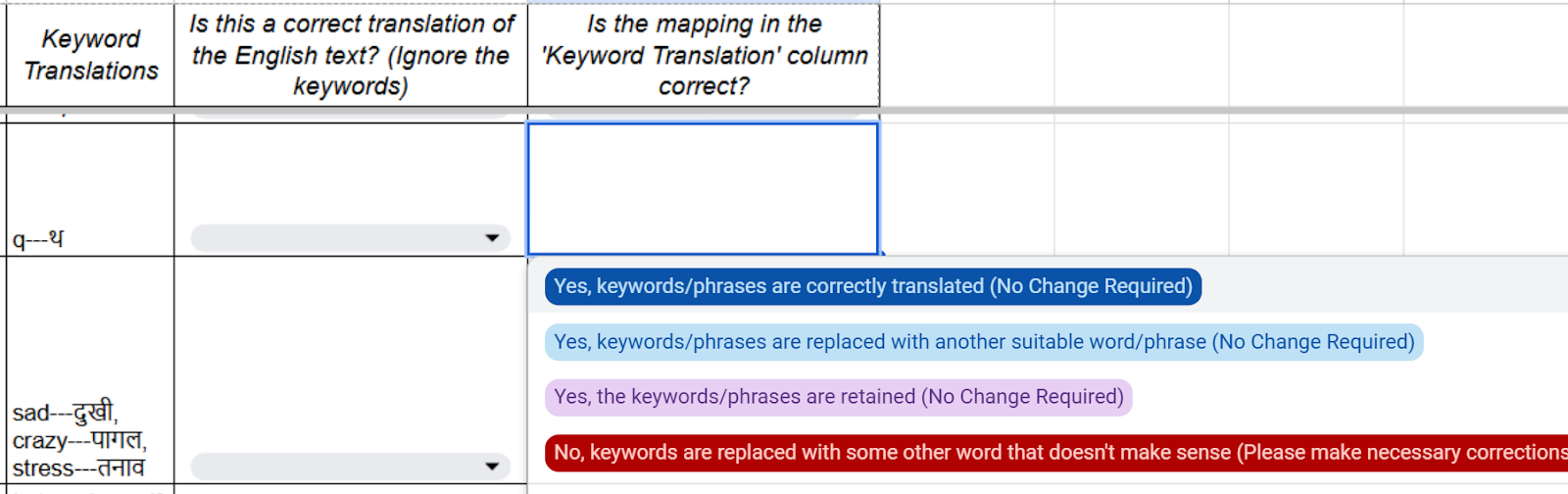}
    \caption{Human Verification for \transset{} conducted on Google Sheets.}
    \label{fig:transset_verification}
\end{figure*}

To summarize, we first collect the list of English constrained keywords from all the prompts. We translate those keywords individually to the target language (``Keyword Translation''). We ensure that there no keyword translation collapse occurs using simple script checks, and if any, rectify them with an alternative keyword using Gemini-2.5-Flash. We then insert the translated keywords in the place where the keyword is in the English prompts (``Pretranslation''). We then translate the pretranslations again to get our full prompt translation (``Full Translation''). We replace the keywords back to their original forms where the response is expected in another language (e.g., English). We then perform a final check using a regex checks to see whether the keywords that are inserted in the pretranslation are actually preserved after full translation, and if not, we manually insert them to the respective positions. Though this method does not avoid manual modification entirely after translation, it substantially reduces the amount of corrective editing required, especially in cases where targeted machine translation systems are available.

Most importantly, we believe this comprehensive pipeline offloads mechanically verifiable checks wherever possible to automated scripts, allowing annotators to focus more on assessing semantic correctness of the translations.

We also perform many minor modifications in general that would ease automatic translation. For example, the keyword `cook' can be interpreted as a chef, or the act of cooking. When translated without context, this could result in a mistranslation. In the particular example where this was the case, the contextless translation rendered it as `chef', and so to avoid this, we manually modified `cook' into `cooking'. Thus, as much as possible, we modify the English keywords to minimize such errors during keyword translation phase, reducing the required time and effort of the annotators.

\section{Grounding}
\label{sec:appendix:grounding} 
To ensure the keywords used were culturally and linguistically relevant, we mined high-value terms from the Sangraha corpus using TF-IDF ranking, selecting approximately 30 keywords per language. Using these terms, we retrieved roughly 300–500 existing text segments per language from the corpus that naturally satisfied a specific constraint, such as a keyword appearing a set number of times. Finally, we provided these naturally occurring contexts along with their satisfied constraints to Gemini-2.5-Pro, which generated natural language prompts that would plausibly lead to the retrieved response, yielding a set of naturally grounded instruction-response pairs.

\begin{figure*}[h]
    \centering
    \includegraphics[width=1.0\linewidth]{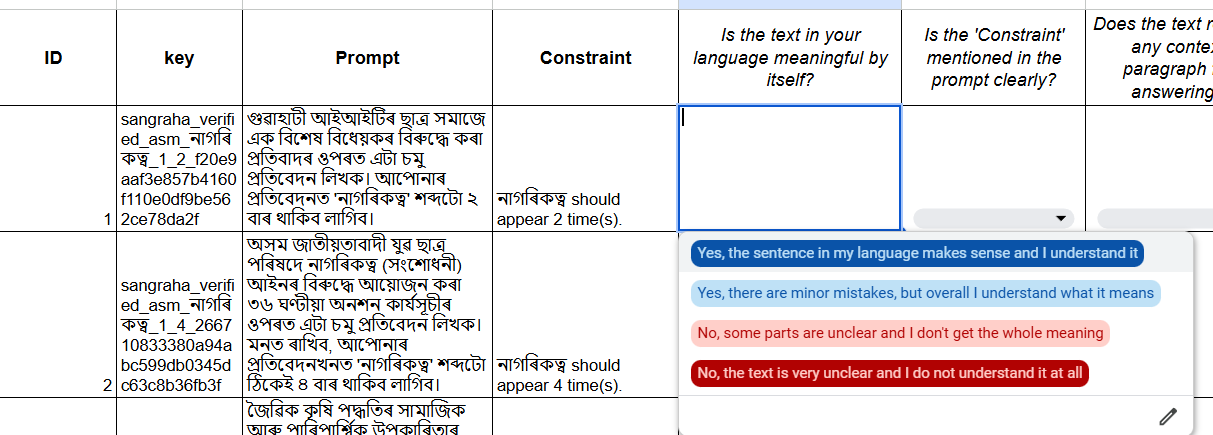}
    \includegraphics[width=1.0\linewidth]{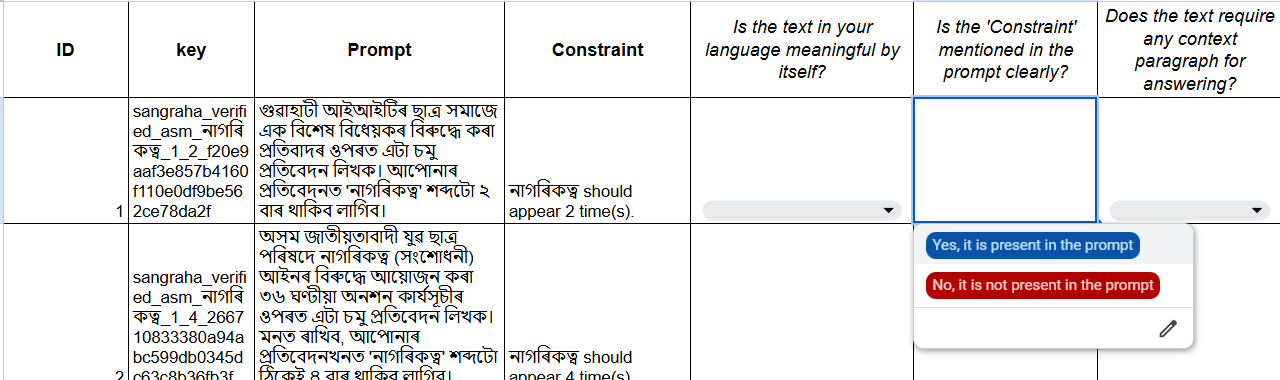}
    \caption{Human Verification for \groundset{} conducted on Google Sheets.}
    \label{fig:groundset_verification}
\end{figure*}

\begin{figure}[h]
    \centering
    \includegraphics[width=1.0\linewidth]{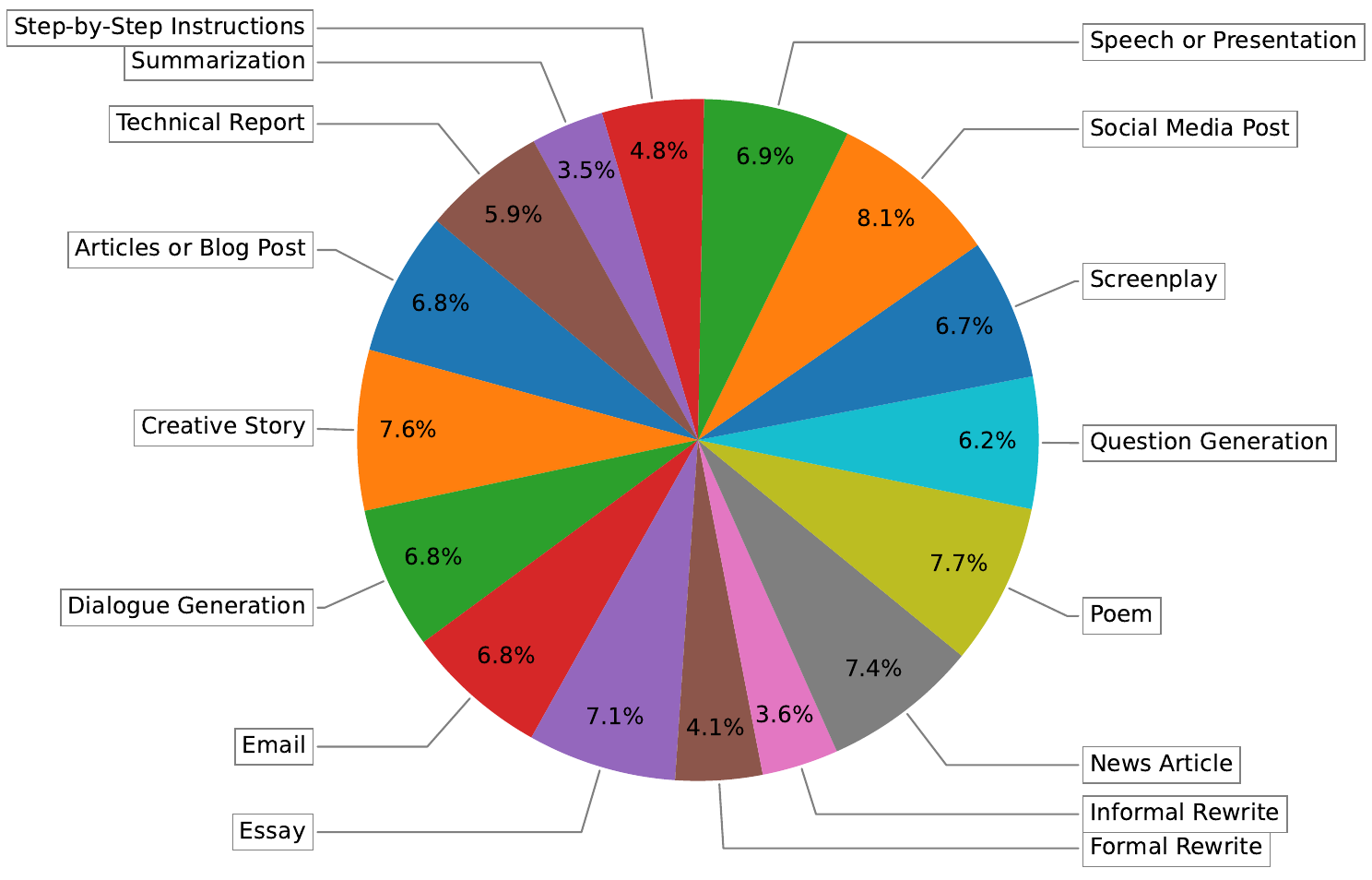}
    \caption{Proportion of the instruction tasks in \groundset{}, averaged across languages.}
    \label{fig:groundset_task_proportions}
\end{figure}

To enhance instruction diversity during this prompt generation phase, we instructed the model to format the output according to one of sixteen distinct text generation tasks inspired by the UltraChat framework \cite{ding2023enhancing}. We detail these tasks in Table \ref{tab:ultrachat_tasks} and illustrate the average proportion of each task across the dataset in Figure \ref{fig:groundset_task_proportions}.

We provide the prompts used for generating these instructions in Figure \ref{fig:prompt_template_1}, \ref{fig:prompt_template_2}, \ref{fig:prompt_template_3}, and \ref{fig:prompt_template_4}. 

\begin{table}[h]
\centering
\small
\begin{tabular}{ll}
\toprule
\multicolumn{2}{c}{\textbf{\groundset{} Instruction Tasks}} \\
\midrule
\cellcolor{openbest} News Article & \cellcolor{openbest} Step-by-Step Instructions \\
\cellcolor{openbest} Articles or Blog Post & \cellcolor{openbest} Social Media Post \\
\cellcolor{openbest} Essay & \cellcolor{openbest} Technical Report \\
\cellcolor{openbest} Poem & \cellcolor{openbest} Question Generation \\
\cellcolor{openbest} Creative Story & \cellcolor{openbest} Email \\
\cellcolor{openbest} Dialogue Generation & \cellcolor{propbest} Summarization \\
\cellcolor{openbest} Screenplay & \cellcolor{propbest} Formal Rewrite \\
\cellcolor{openbest} Speech or Presentation & \cellcolor{propbest} Informal Rewrite \\
\bottomrule
\end{tabular}
\caption{List of generation tasks used for prompt diversification of the \groundset{} dataset. Red color indicates instructions that require the context be included in the generated prompt; blue represents tasks that require no context for answering.}
\label{tab:ultrachat_tasks}
\end{table}

\begin{table}[!h]
\centering
\small
\setlength{\tabcolsep}{8pt}
\renewcommand{\arraystretch}{1.2}

\begin{tabular}{|l|c|}
\hline
\textbf{Language} & \textbf{\# Correct Prompts} \\
\hline
Assamese & 257/439 \\
Bengali & 409/502 \\
Gujarati & 377/446 \\
Hindi & 457/527 \\
Kannada & 341/392 \\
Malayalam & 350/446 \\
Marathi & 384/415 \\
Nepali & 341/419 \\
Odia & 368/460 \\
Punjabi & 409/414 \\
Sanskrit & 336/406 \\
Tamil & 430/442 \\
Telugu & 375/430 \\
Urdu & 337/399 \\
\hline
\end{tabular}

\caption{Per-language prompt verification statistics for \groundset{}.}
\label{tab:synthetic-verification-stats}
\end{table}

\label{sec:appendix:verification}

\section{Human Verification}
For each language, we employ one annotator who is a native speaker with linguistic qualifications. 

While multi-annotator validation is the established standard for dataset reliability, our pipeline relies on a single native speaker per language due to funding constraints. To maintain high quality within these limits, we enforce strict verification guidelines to eliminate subjective editing. Annotators must either accept a prompt or drop it entirely from the dataset. Prompts with minor grammatical mistakes are accepted if the overall meaning is fully understandable, as our evaluation focuses strictly on adherence to the specified constraints. We acknowledge the absence of inter-annotator agreement metrics as a limitation and plan to incorporate multi-annotator spot checks in future iterations.

\subsection{\transset{}}

\begin{table}[!h]
\centering
\small
\setlength{\tabcolsep}{6pt}
\renewcommand{\arraystretch}{1.2}

\begin{tabular}{|l|>{\centering\arraybackslash}p{2cm}|>{\centering\arraybackslash}p{2cm}|}
\hline
\textbf{Language} &
\centering\textbf{\# Correct Translations (Total 490)}\arraybackslash &
\centering\textbf{\# Keyword Corrections (Total 200)}\arraybackslash \\
\hline
Assamese   & 440 & 38 \\
Bengali    & 454 & 21 \\
Gujarati   & 482 & 8 \\
Hindi      & 487 & 9 \\
Kannada    & 457 & 27 \\
Malayalam  & 490 & 5 \\
Marathi    & 483 & 35 \\
Nepali     & 468 & 58 \\
Odia       & 473 & 19 \\
Punjabi    & 474 & 16 \\
Sanskrit   & 450 & 61 \\
Tamil      & 489 & 32 \\
Telugu     & 488 & 34 \\
Urdu       & 486 & 5 \\
\hline
\end{tabular}
\caption{Per-language prompt verification statistics for \transset{}. Parallely across languages, there are 321 correct translations.}
\label{tab:translated-verification-stats}
\end{table}

To ensure the quality and reliability of \transset{}, we conducted a structured verification process involving native speakers of the target languages. The goal of this process was to assess both the semantic correctness of the translations and the preservation of instruction-level constraints present in the original English IFEval prompts.

The translated dataset was verified using a set of annotation guidelines shared with all annotators prior to the verification task. Annotators were instructed to evaluate each translated prompt independently, without rewriting or paraphrasing the text. The verification focused on whether the translated prompt was meaningful as a standalone instruction in the target language and whether it faithfully reflected the intent, constraints, and content of the original English prompt.

The verification was carried out using two separate annotation workflows, corresponding to prompts without keyword constraints and prompts with explicit keyword or phrase constraints. For prompts without keywords, annotators assessed (i) whether the translated prompt was meaningful and (ii) whether it constituted a correct translation of the English source. For prompts with keyword constraints, annotators additionally verified the correctness and contextual suitability of the translated keywords, following clearly defined rules regarding named entities, language-specific adaptations, and constraint preservation.

Importantly, annotators were not required to manually edit or rewrite translations, except in rare cases where keyword mappings were clearly incorrect. All other checks were limited to selecting predefined annotation options. This design allowed mechanically verifiable checks to be handled systematically while enabling annotators to focus primarily on semantic correctness.

Following this verification process, prompts marked as incorrect were dropped, and in cases where prompt was marked correct, but keywords were incorrectly translated, they were simply updated in the prompt. The final validated dataset statistics are presented in Table \ref{tab:translated-verification-stats}

\subsection{\groundset{}}

The Indic-grounded prompts were automatically generated and paired with explicit constraint descriptions, such as requirements on keyword usage or sentence count.

Annotators were provided with detailed guidelines and evaluated each synthetic prompt using a structured annotation interface. For every prompt, annotators assessed three aspects: (i) whether the prompt is meaningful and natural in the target language, (ii) whether the stated constraint is clearly and correctly expressed in the prompt, and (iii) whether the prompt requires an external context paragraph to be answerable.

Annotators were instructed not to rewrite or improve the prompts. Instead, they selected predefined responses indicating the presence or absence of errors. This design ensures that verification focuses on correctness and clarity rather than stylistic preference. Prompts identified as unclear, incorrect, or missing constraints were excluded.
\newpage

\begin{figure}
\begin{tcolorbox}[colback=propbest, colframe=propbest, fontupper=\ttfamily\small, arc=4pt, boxrule=0pt]
Role: You are an expert prompt engineer specializing in creating evaluation prompts for Large Language Models (LLMs). Your task is to design a high-quality, creative prompt for the IFEval benchmark (in \{\{LANGUAGE\}\}), which is designed to test an LLM's ability to follow specific instructions and constraints.\\

Objective: I will provide you with a Constraint and an Example Context (in \{\{LANGUAGE\}\}). Based on this input, you will generate a new prompt that instructs an LLM to perform a task while adhering to the given constraint.\\

Strict Requirement: \{\{INSTRUCTION TASK\}\}\\

[START OF INPUT]

Constraint to Test:
\{\{KEYWORD\}\} should appear \{\{FREQUENCY\}\} time(s).

[END OF INPUT]\\

Example Context (for grounding/theme):
\{\{CONTEXT\}\}
[END OF INPUT]\\

Your Task:
Based on the Constraint to Test and the Example Context provided above, generate a single, new prompt in JSON format.\\

[START OF OUTPUT]

```json

\{
    "instruction": "[generated instruction with mention of constraint]",
    "constraint": "\{\{KEYWORD\}\} should appear \{\{FREQUENCY\}\} time(s)."
\}

```

[END OF OUTPUT]\\

Requirements for the Generated Prompt:
- Natural Integration: The constraint must be woven seamlessly and logically into the prompt's instructions.\\

- Clear Task: The prompt must define a clear and achievable task for the LLM.

- Creative Context: The prompt should be engaging and provide a context that makes the constraint feel purposeful rather than arbitrary.

- No Meta-Language: Do not mention "LLM," "AI," "test," "evaluation," or "IFEval." The prompt should feel like a genuine request.

- Output: Provide only the generated prompt without any additional explanation or commentary in the form of json format containing two keys:

    1) `instruction`: The generated instruction string.
    
    2) `constraint`: The constraint description string.
    
- Safety: The generated instruction should be safe to use. If the context itself is unsafe or controversial, please generate "This context contains controversial themes" only.
\end{tcolorbox}
\caption{Prompt template used to generate instructions for Keyword Inclusion \& Frequency.}
\label{fig:prompt_template_1}
\end{figure}

\begin{figure}
\begin{tcolorbox}[colback=openbest, colframe=openbest, fontupper=\ttfamily\small, arc=4pt, boxrule=0pt]
Role: You are an expert prompt engineer specializing in creating evaluation prompts for Large Language Models (LLMs). Your task is to design a high-quality, creative prompt for the IFEval benchmark (in \{\{LANGUAGE\}\}), which is designed to test an LLM's ability to follow specific instructions and constraints.\\

Objective: I will provide you with a Constraint and an Example Context (in \{\{LANGUAGE\}\}). Based on this input, you will generate a new prompt that instructs an LLM to perform a task while adhering to the given constraint.\\

Strict Requirement: \{\{INSTRUCTION TASK\}\}\\

[START OF INPUT]

Constraint to Test:
\{\{KEYWORD\}\} should be the first word in the response.

[END OF INPUT]\\

Example Context (for grounding/theme):
\{\{CONTEXT\}\}
[END OF INPUT]\\

Your Task:
Based on the Constraint to Test and the Example Context provided above, generate a single, new prompt in JSON format.\\

[START OF OUTPUT]

```json

\{
    "instruction": "[generated instruction with mention of constraint]",
    "constraint": "\{\{KEYWORD\}\} should be the first word in the response."
\}

```

[END OF OUTPUT]\\

Requirements for the Generated Prompt:
- Natural Integration: The constraint must be woven seamlessly and logically into the prompt's instructions.\\

- Clear Task: The prompt must define a clear and achievable task for the LLM.

- Creative Context: The prompt should be engaging and provide a context that makes the constraint feel purposeful rather than arbitrary.

- No Meta-Language: Do not mention "LLM," "AI," "test," "evaluation," or "IFEval." The prompt should feel like a genuine request.

- Output: Provide only the generated prompt without any additional explanation or commentary in the form of json format containing two keys:

    1) `instruction`: The generated instruction string.
    
    2) `constraint`: The constraint description string.
    
- Safety: The generated instruction should be safe to use. If the context itself is unsafe or controversial, please generate "This context contains controversial themes" only.
\end{tcolorbox}
\caption{Prompt template used to generate instructions for First Word constraints.}
\label{fig:prompt_template_2}
\end{figure}

\begin{figure}
\begin{tcolorbox}[colback=propbest, colframe=propbest, fontupper=\ttfamily\small, arc=4pt, boxrule=0pt]
Role: You are an expert prompt engineer specializing in creating evaluation prompts for Large Language Models (LLMs). Your task is to design a high-quality, creative prompt for the IFEval benchmark (in \{\{LANGUAGE\}\}), which is designed to test an LLM's ability to follow specific instructions and constraints.\\

Objective: I will provide you with a Constraint and an Example Context (in \{\{LANGUAGE\}\}). Based on this input, you will generate a new prompt that instructs an LLM to perform a task while adhering to the given constraint.\\

Strict Requirement: \{\{INSTRUCTION TASK\}\}\\

[START OF INPUT]

Constraint to Test:

The response should have \{\{SENTENCES\}\}  sentence(s) and \{\{PARAGRAPHS\}\} "paragraph(s).

[END OF INPUT]\\

Example Context (for grounding/theme):
\{\{CONTEXT\}\}
[END OF INPUT]\\

Your Task:
Based on the Constraint to Test and the Example Context provided above, generate a single, new prompt in JSON format.\\

[START OF OUTPUT]

```json

\{
    "instruction": "[generated instruction with mention of constraint]",
    "constraint": "The response should have \{\{SENTENCES\}\}  sentence(s) and \{\{PARAGRAPHS\}\} "paragraph(s)."
\}

```

[END OF OUTPUT]\\

Requirements for the Generated Prompt:
- Natural Integration: The constraint must be woven seamlessly and logically into the prompt's instructions.\\

- Clear Task: The prompt must define a clear and achievable task for the LLM.

- Creative Context: The prompt should be engaging and provide a context that makes the constraint feel purposeful rather than arbitrary.

- No Meta-Language: Do not mention "LLM," "AI," "test," "evaluation," or "IFEval." The prompt should feel like a genuine request.

- Output: Provide only the generated prompt without any additional explanation or commentary in the form of json format containing two keys:

    1) `instruction`: The generated instruction string.
    
    2) `constraint`: The constraint description string.
    
- Safety: The generated instruction should be safe to use. If the context itself is unsafe or controversial, please generate "This context contains controversial themes" only.
\end{tcolorbox}
\caption{Prompt template used to generate instructions for Sentence \& Paragraph Count constraints.}
\label{fig:prompt_template_3}
\end{figure}

\begin{figure}
\begin{tcolorbox}[colback=openbest, colframe=openbest, fontupper=\ttfamily\small, arc=4pt, boxrule=0pt]
Role: You are an expert prompt engineer specializing in creating evaluation prompts for Large Language Models (LLMs). Your task is to design a high-quality, creative prompt for the IFEval benchmark (in \{\{LANGUAGE\}\}), which is designed to test an LLM's ability to follow specific instructions and constraints.\\

Objective: I will provide you with a Constraint and an Example Context (in \{\{LANGUAGE\}\}). Based on this input, you will generate a new prompt that instructs an LLM to perform a task while adhering to the given constraint.\\

Strict Requirement: \{\{INSTRUCTION TASK\}\}\\

[START OF INPUT]

Constraint to Test:

\{\{KEYWORD\}\} should not appear in the response."
[END OF INPUT]\\

Example Context (for grounding/theme):
\{\{CONTEXT\}\}
[END OF INPUT]\\

Your Task:
Based on the Constraint to Test and the Example Context provided above, generate a single, new prompt in JSON format.\\

[START OF OUTPUT]

```json

\{
    "instruction": "[generated instruction with mention of constraint]",
    "constraint": "\{\{KEYWORD\}\} should not appear in the response."
\}

```

[END OF OUTPUT]\\

Requirements for the Generated Prompt:
- Natural Integration: The constraint must be woven seamlessly and logically into the prompt's instructions.\\

- Clear Task: The prompt must define a clear and achievable task for the LLM.

- Creative Context: The prompt should be engaging and provide a context that makes the constraint feel purposeful rather than arbitrary.

- No Meta-Language: Do not mention "LLM," "AI," "test," "evaluation," or "IFEval." The prompt should feel like a genuine request.

- Output: Provide only the generated prompt without any additional explanation or commentary in the form of json format containing two keys:

    1) `instruction`: The generated instruction string.
    
    2) `constraint`: The constraint description string.
    
- Safety: The generated instruction should be safe to use. If the context itself is unsafe or controversial, please generate "This context contains controversial themes" only.
\end{tcolorbox}
\caption{Prompt template used to generate instructions Forbidden Keyword constraints.}
\label{fig:prompt_template_4}
\end{figure}

\begin{table*}[t]
    \centering
    \includegraphics[width=\linewidth]{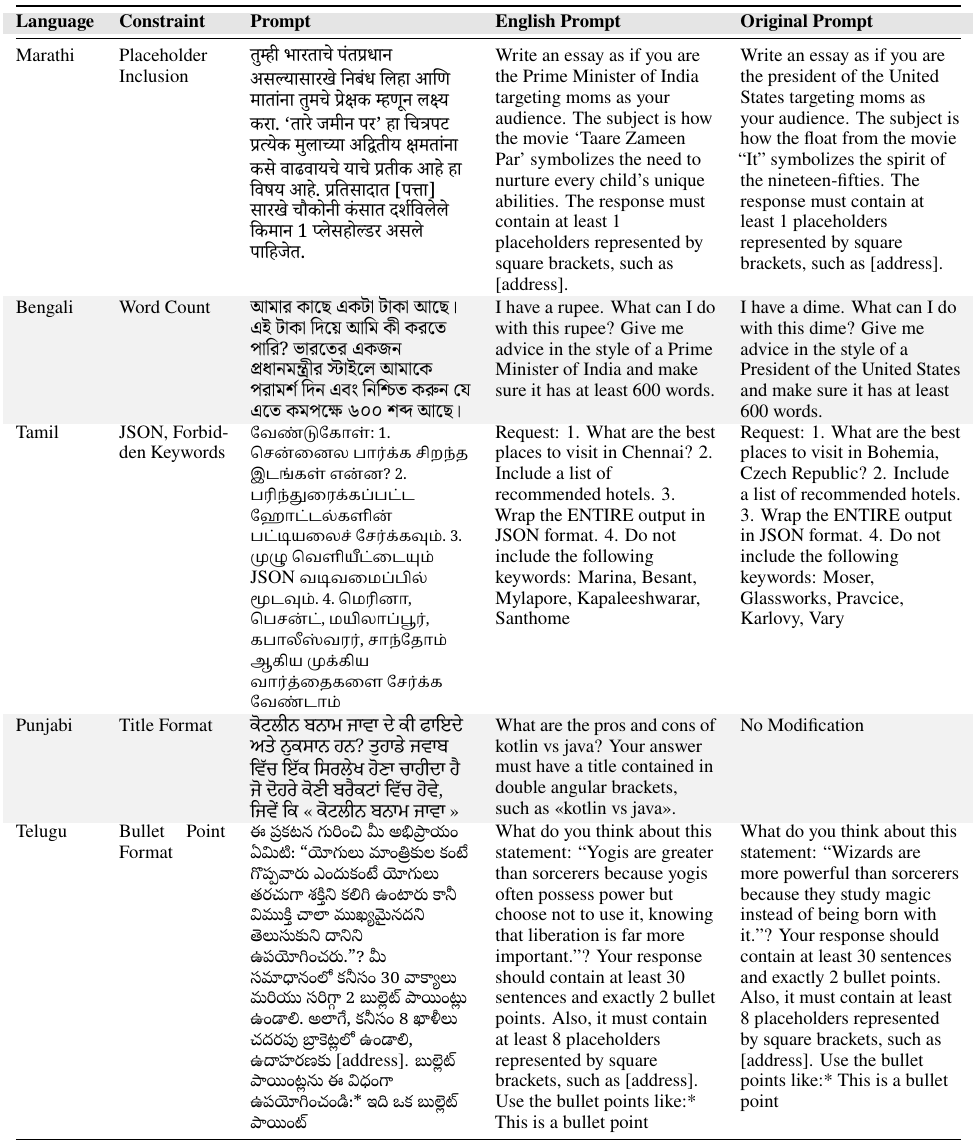}
    \caption{Examples of prompt-constraint pairs from the \transset{} dataset. The final column contains the original, unmodified prompt from the IFEval dataset \cite{ifeval}.}
    \label{tab:translated_examples}
\end{table*}
 
\begin{table*}[t]
    \centering
    \includegraphics[width=\linewidth]{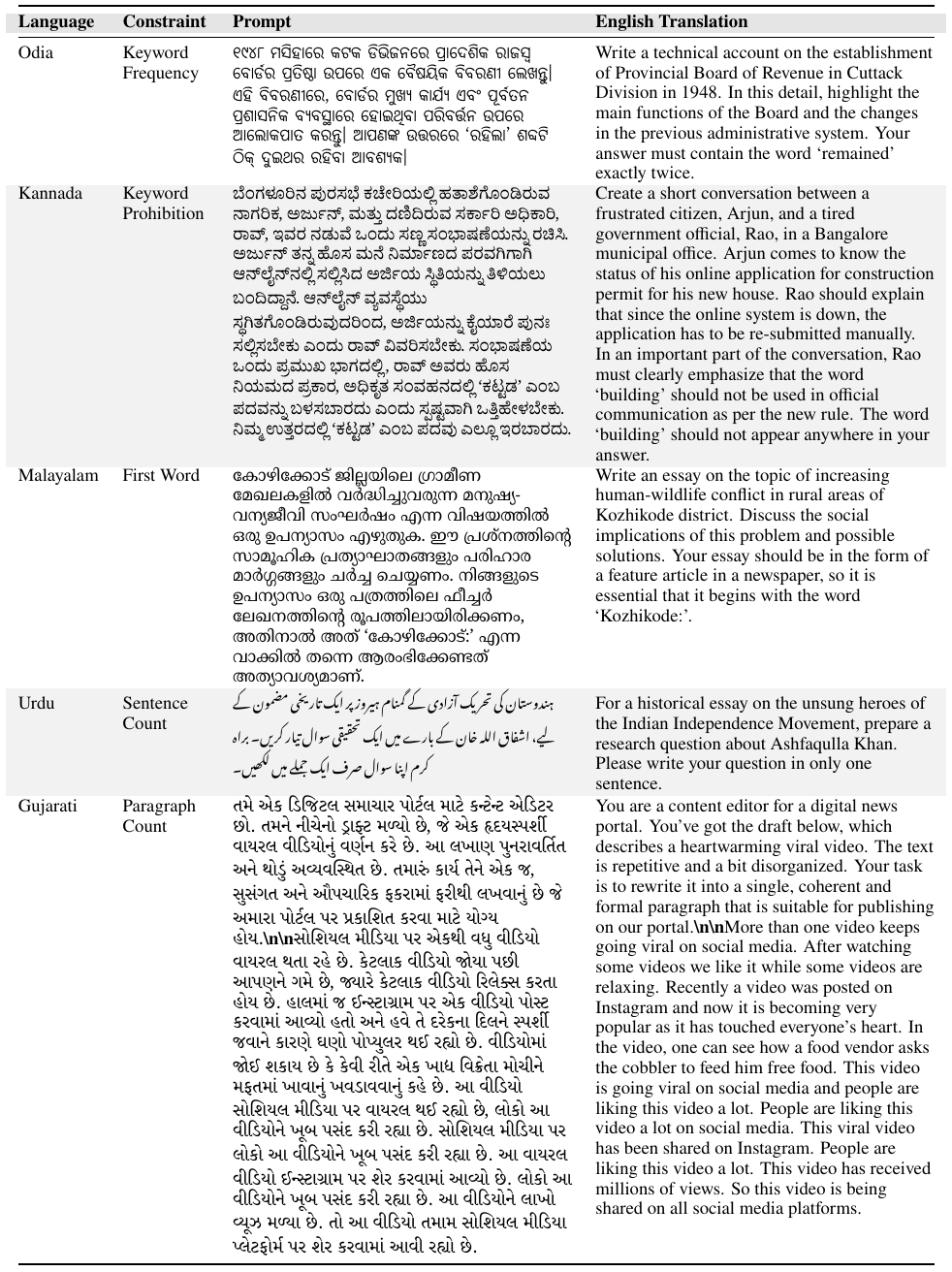}
    \caption{Examples of prompt-constraint pairs from the \groundset{} dataset.}
    \label{tab:grounded_examples}
\end{table*}

\end{document}